\documentclass{article}

\usepackage{arxiv}
\newcommand{\PaperTitle}{Spherical Cauchy Variational Autoencoders: Heavy Angular Tails and Exact KL Evaluation}
\newcommand{\PaperShortTitle}{Spherical Cauchy Latent Variables}
\newcommand{\PaperAuthors}{Lukas Sablica and Kurt Hornik}
\newcommand{\PaperShortAuthors}{Sablica and Hornik}
\newcommand{\PaperInstitution}{Institute for Statistics and Mathematics}
\newcommand{\PaperUniversity}{Vienna University of Economics and Business}
\newcommand{\PaperCountry}{Austria}
\newcommand{\LukasEmail}{Lukas.Sablica@wu.ac.at}
\newcommand{\LukasORCID}{0000-0001-9166-4563}
\newcommand{\KurtEmail}{Kurt.Hornik@wu.ac.at}
\newcommand{\KurtORCID}{0000-0003-4198-9911}
\newcommand{\PaperKeywords}{variational autoencoders, hyperspherical latent variables, spherical Cauchy distribution,
heavy-tailed posteriors, M\"obius reparameterization, Kullback-Leibler divergence, numerical stability}

\newcommand{\PaperFunding}{Lukas Sablica's time during final preparation and submission was supported by funds of the Oesterreichische Nationalbank, Austrian Central Bank Anniversary Fund, project number 19079. The research underlying the manuscript was conducted before this funding.}

\usepackage[utf8]{inputenc}
\usepackage[T1]{fontenc}
\usepackage{amsmath,amssymb,amsfonts,mathtools}
\usepackage{bm}
\usepackage{booktabs,array}
\usepackage{graphicx}
\usepackage{float}
\usepackage{microtype}
\usepackage{url}

\newcommand{\SC}{\operatorname{spCauchy}}
\newcommand{\vMF}{\operatorname{vMF}}
\newcommand{\PS}{\operatorname{PowerSpherical}}
\newcommand{\KL}{\operatorname{KL}}
\newcommand{\SD}{\mathbb{S}^{D-1}}
\newcommand{\BD}{\mathbb{B}^{D}}
\newcommand{\R}{\mathbb{R}}
\newcommand{\E}{\mathbb{E}}
\newcommand{\nuD}{\upsilon_D}
\newcommand{\ms}[2]{\ensuremath{#1\,{\scriptscriptstyle\pm}\,#2}}
\newcommand{\bms}[2]{\ensuremath{\boldsymbol{#1\,{\scriptscriptstyle\pm}\,#2}}}
\newcommand{\bmsd}[2]{\ensuremath{\boldsymbol{#1\,{\scriptscriptstyle\pm}\,#2}^{\dagger}}}

\usepackage{amsthm}
\usepackage[round]{natbib}
\usepackage[colorlinks=true,linkcolor=blue,citecolor=blue,urlcolor=blue,hypertexnames=false]{hyperref}

\newcommand{\SupportingMaterialName}{Appendix}
\newcommand{\supportingMaterialName}{appendix}
\newcommand{\PaperCitep}[1]{\citep{#1}}
\newcommand{\PaperCitet}[2]{\citet{#2}}
\newif\ifSCVAEJournalVersion
\SCVAEJournalVersionfalse

\newtheorem{theorem}{Theorem}
\newtheorem{lemma}[theorem]{Lemma}
\newtheorem{proposition}[theorem]{Proposition}
\newtheorem{corollary}[theorem]{Corollary}
\newtheorem{suppprop}{Proposition}

\renewcommand{\headeright}{\PaperShortAuthors}
\renewcommand{\undertitle}{A Preprint}
\title{\PaperTitle}
\author{
  Lukas Sablica \\
  \PaperInstitution\\
  \PaperUniversity\\
  \PaperCountry\\
  ORCiD: \LukasORCID\\
  \href{mailto:\LukasEmail}{\nolinkurl{\LukasEmail}}
  \And
  Kurt Hornik\\
  \PaperInstitution\\
  \PaperUniversity\\
  \PaperCountry\\
  ORCiD: \KurtORCID\\
  \href{mailto:\KurtEmail}{\nolinkurl{\KurtEmail}}
}
\date{}

\hypersetup{
  pdftitle={\PaperTitle},
  pdfauthor={\PaperAuthors}
}

\begin{document}

\maketitle
\fancyhead[C]{\PaperShortTitle}

\begin{abstract}
Heavy-tailed posteriors are routine in Euclidean variational autoencoders, where the Student family relaxes the Gaussian without new machinery. The sphere has had no comparable option. Von Mises-Fisher distribution needs modified Bessel functions and a rejection sampler, and Power Spherical buys its closed forms by forcing the density to vanish at the antipode. We develop the spherical Cauchy distribution as a hyperspherical posterior that needs neither compromise. Stereographic projection carries it to a multivariate Student law, and a M\"obius transformation turns a uniform spherical draw into an exact posterior sample from inner products, norms, and scalar arithmetic. The same transformation settles the regularizer. Evaluating the density along the sampling map reduces the Kullback-Leibler (KL) divergence to the uniform prior to a scalar expectation whose expansion terminates in every even ambient dimension, leaving one logarithm and a polynomial with finitely many correction terms. Odd dimensions admit certified truncation of value and gradient, the KL is increasing and convex in concentration, and the same function gives the pairwise KL. At matched modal curvature it has broader angular tails and a smaller KL penalty than both alternatives, so equal local precision costs less regularization. In dimension 128 the fused evaluator runs 1.5 times faster per latent-layer step than Power Spherical and 4.2 times faster than robust von Mises-Fisher on CPU, with factors of 1.6 and 5.4 on CUDA. Across five paired seeds it attains the lowest MNIST reconstruction loss at every tested dimension and lowers held-out viewpoint-gap negative log-likelihood on smallNORB by 3.6 percent.

\end{abstract}


\section{Introduction}
\label{sec:introduction}

The geometry of a latent space determines which representations a generative model treats as close and how it expresses uncertainty between them. For directional features and periodic factors such as object azimuth, angular relationships can be more meaningful than distances in an unconstrained Euclidean space. A circle, for example, keeps viewpoints just below and just above $0^\circ$ close without introducing an artificial boundary. Normalized embeddings and directional training objectives exploit related geometric structure in learned representations \PaperCitep{scott2021mises,scott2022empirical}.

Hyperspherical variational autoencoders make these angular relationships part of the probabilistic model \PaperCitep{davidson2018hyperspherical,xu2018spherical}. Latent codes have a fixed norm, and a uniform spherical prior gives every direction equal status. The posterior can therefore express a preferred direction and uncertainty around it without also learning a radial scale. This is useful when magnitude is incidental to the representation, and recent generative models place such spherical bottlenecks at the center of the model \PaperCitep{ke2025spherear,bond2026s2vae}. Making that choice practical, however, requires a posterior that can be sampled and regularized efficiently throughout training.

The posterior also determines how much probability is retained far from the preferred direction. In Euclidean variational autoencoders \PaperCitep{VAEs}, a Student posterior can replace a Gaussian one while retaining the encoder-decoder framework \PaperCitep{kim2024t3vae}. An equally convenient heavy-tailed option would bring this flexibility to spherical models while preserving precise local information. The two spherical posteriors in regular use each put a price on this choice.

The von Mises-Fisher (vMF) posterior is the canonical rotationally symmetric exponential family on the sphere, and its statistical credentials are not in question \PaperCitep{davidson2018hyperspherical,xu2018spherical}. Its price is numerical, and it is charged twice. Pathwise sampling goes through rejection \PaperCitep{Ulrich}, and an acceptance step is not differentiable, so the reparameterization trick has to be repaired by the correction machinery of reparameterized rejection sampling \PaperCitep{Naesseth}. The normalizer, entropy, moments, and KL divergence are then built from modified Bessel functions whose order moves with the ambient dimension. Stable evaluation of these ratios is a solved problem \PaperCitep{hornik2014maximum,movMF}, but the solutions are iterative loops and branch-heavy scalar code, which is what a GPU handles worst.

Power Spherical (PS) was introduced to remove precisely this machinery, and it does \PaperCitep{decao2020power}. A power of the shifted cosine gives a Beta marginal, an explicit sampler, and a KL divergence in Gamma and digamma functions. The construction is design rather than derivation. It also commits the family to a global shape. For every positive exponent the density is exactly zero at the antipode, and its angular score diverges as the opposite direction is approached. Where the opposite direction is genuinely impossible this is a feature. For a generic latent code it is an assumption nobody asked for. Neither price has to be paid.

We study the spherical Cauchy family of \PaperCitet{Kato and McCullagh}{spCauchy}. A mean direction $\mu$ and a concentration $\rho\in[0,1)$ index a density that is one power of the squared distance to the point $a=\rho\mu$ of the open unit ball. We work with $a$ directly, since it carries both parameters at once and leaves no undefined direction at $\rho=0$. The density is strictly positive at every angle, and stereographic projection carries it to a multivariate Student law whose degrees of freedom equal the intrinsic dimension of the sphere. It can also be sampled directly, since a M\"obius transformation carries the uniform sphere onto it. An exact pathwise sample is a normalized Gaussian draw followed by one rational map, with no acceptance step and no implicit gradient. That combination is what recommends the family as a variational posterior. It offers what the Student law offers in the Euclidean case, and it offers it with a sampler the reparameterization trick can use unmodified.

This supplies the sampling part of a VAE, but training also requires a tractable regularizer. The reconstruction term is balanced against a KL penalty that limits how much information each posterior carries relative to the prior. Both terms must provide reliable gradients to the encoder, and the KL must be evaluated for every observation at every step. Our central contribution is to make this second component tractable for spherical Cauchy using the same transport that supplies the sampler. Evaluating the density along the map that generates it collapses the KL divergence to the uniform prior into a scalar expectation over the uniform sphere. The remaining questions are how to evaluate this expectation with controlled cost and accuracy, and when the resulting posterior is preferable to the existing spherical alternatives.

We address the evaluation through a Fourier expansion in the polar angle. Odd moments vanish by symmetry, the even moments of the uniform spherical angle are a ratio of Pochhammer symbols, and the result is
\begin{equation}
\label{eq:intro-kl}
K_D(\rho)
:=\KL\!\left(\SC_D(\mu,\rho)\,\middle\|\,\nuD\right)
=(D-1)\left[
-\log(1-\rho^2)
-\sum_{j=1}^{\infty}
\frac{(1-D/2)_j}{(D/2)_j}\frac{\rho^{2j}}{j}
\right],
\end{equation}
where $\nuD$ denotes the uniform probability measure on $\mathbb{S}^{D-1}$. The logarithm carries the entire divergence as $\rho\to1^-$, and the correction after it is bounded and absolutely convergent up to that boundary. If $D=2q$ the numerator is $(1-q)_j$ and vanishes once $j$ reaches $q$, so the correction is a polynomial in $\rho^2$ with $q-1$ terms. Powers of two and most latent dimensions in ordinary use are exact cases.

Expanding the logarithm and collecting equal powers leaves only positive coefficients, so the KL increases and is convex in the concentration, and its gradient comes for free from the same numbers. Odd dimensions do not terminate, but they certify. Value and gradient each come with a computable bound on the truncation error, so a requested accuracy is met at any concentration and no separate high-concentration branch is needed. A finite alternative built from the two neighboring even dimensions stays within $10^{-3}$ of the exact value when the certificate is not required. The same function also returns the KL between two spherical Cauchy laws rather than only between one and the uniform prior, which extends the evaluator to learned and conditional priors at no cost.

That settles the cost. The second question is what the family buys, and the comparison is only meaningful at fixed local precision. We match the quadratic curvature of the log density at the mode across the three families, which pins down how sharp each posterior is and leaves only the global shape free. The ordering is then strict and complete. Power Spherical has the lightest angular tails and the largest KL to the uniform prior, von Mises-Fisher lies between, and spherical Cauchy has the heaviest tails and the smallest KL. The angular likelihood ratio is monotone between neighboring pairs, so the ordering holds at every radius and for every nondecreasing function of the angle. Spherical Cauchy therefore keeps the same local precision while claiming much less about everything else, which is the weakly informative principle of Bayesian modeling arriving on the posterior side. Heavy tails are chosen there to be sharp where the information is without declaring remote values impossible, and the ordering shows that on the sphere this is not only a matter of taste.

Exactness turns out not to cost speed. A fused evaluator for value and gradient runs faster per latent-layer step than a numerically robust von Mises-Fisher implementation, and faster than Power Spherical as well, on CPU and on CUDA. The second comparison is the informative one, since Power Spherical gave up its antipodal support in exchange for being quick. We then run a controlled test that isolates the geometry from the arithmetic. The targets are von Mises-Fisher densities mixed with a small uniform component, and each family is fitted to them in turn. With no uniform component at all, von Mises-Fisher fits best, as it must. One percent of uniform mass is enough to make spherical Cauchy the better fit, and it stays the best fit as the contamination grows. The end-to-end results follow the same pattern, with the lowest held-out reconstruction loss at every latent dimension tested on MNIST \PaperCitep{lecun1998mnist} and in every seed across a held-out viewpoint sector on smallNORB \PaperCitep{lecun2004norb}.

The distribution and its transport are classical. What we contribute is the latent layer built from them, the exact KL theory that the same transport generates, the matched-curvature comparison against the two spherical posteriors this one would replace, evaluation rules with error certificates in every dimension and parity, and an implementation tested from the latent layer through to reconstruction. All proofs are in the \SupportingMaterialName.

\section{Related work}
\label{sec:related}

Variational autoencoders optimize a stochastic lower bound through a reparameterized posterior \PaperCitep{VAEs}, and the Gaussian posterior is still the default. Attempts to loosen it have gone in three directions, namely richer priors \PaperCitep{tomczak2018vampprior}, alternative divergences \PaperCitep{daudel2023alpha}, and heavy-tailed Euclidean families \PaperCitep{kim2024t3vae}. The last is closest to what we do. The difference is that we stay on the sphere, hold the prior fixed and uniform, and change only the posterior and the exact regularizer that comes with it.

On the sphere the starting point is vMF, which gives an intrinsic latent space and reduces posterior collapse in low-dimensional and text models \PaperCitep{davidson2018hyperspherical,xu2018spherical}. A follow-up replaces the single sphere with a product of lower-dimensional spheres, each carrying its own concentration, on the grounds that one scalar concentration for the whole sphere becomes a bottleneck as dimension grows \PaperCitep{davidson2019increasing}. That work also notes that the vMF cost against the uniform prior increases with dimension. Proposition~\ref{prop:dimension-monotonicity} makes the same statement exact for spherical Cauchy, in the total KL and in the KL per intrinsic dimension.

Power Spherical was designed for tractability \PaperCitep{decao2020power}. The density $(1+\mu^\top x)^\lambda$ gives a Beta marginal, an explicit sampler, and a KL in Gamma and digamma functions. What deserves a mention here is that the commitment is global rather than local. The density vanishes polynomially at the antipode, with order $2\lambda$, where the same exponent also determines the curvature at the mode. The two behaviors therefore cannot be controlled separately. As shown below, this is an especially strong restriction. Even relative to the already light-tailed vMF law, Power Spherical suppresses remote directions more strongly at matched local curvature. The global shape of Power Spherical is therefore a live design choice in current models. Recent geometry-first encoders use products of Power Spherical factors for strongly compressed hyperspherical bottlenecks \PaperCitep{bond2026s2vae}, while our comparison shows that the same local precision can be achieved without forcing the density to vanish at the antipode.

A different line keeps the sphere and abandons the pointwise divergence. S2WTM aligns the aggregated posterior with a spherical prior under a sliced-Wasserstein distance rather than penalizing each posterior against the prior, again motivated by collapse of the KL term \PaperCitep{adhya2025s2wtm}. Our aim is the opposite. We keep the per-observation KL and make it cheap and exact, so the regularizer stays interpretable as an information cost.

Evidence that the shape of a spherical bottleneck matters at scale is now available outside the VAE literature. Constraining autoregressive image tokens to a fixed-radius sphere through a hyperspherical VAE removes the scale component that destabilizes decoding under classifier-free guidance, and yields state-of-the-art FID among autoregressive generators \PaperCitep{ke2025spherear}. Both that system and the geometry-first encoder above draw their posterior from vMF or Power Spherical, which is the menu this paper extends.

M\"obius-generated Cauchy families on spheres were introduced and studied by \PaperCitet{Kato and McCullagh}{spCauchy}, and have since been used for directional clustering and likelihood-based modeling \PaperCitep{circlus,tsagris2025directional}. Variational inference is where the family has not been taken. We supply the latent layer, the exact uniform and pairwise KL theory, the comparison against the closest spherical alternatives, the evaluation rules, and the generative experiments.

Table~\ref{tab:spherical-families} collects the three rotationally symmetric families used in our analysis. The last column states the ordering proved in Proposition~\ref{prop:profile-order}, which holds after matching quadratic curvature at the mode.

\begin{table}[htbp]
\centering
\small
\setlength{\tabcolsep}{3.5pt}
\renewcommand{\arraystretch}{1.3}

\caption{Structural origin, computational form, and matched-curvature behavior of three spherical posteriors.}
\label{tab:spherical-families}
\begin{tabular}{@{}>{\raggedright\arraybackslash}p{0.14\textwidth}
                    >{\raggedright\arraybackslash}p{0.23\textwidth}
                    >{\raggedright\arraybackslash}p{0.28\textwidth}
                    >{\raggedright\arraybackslash}p{0.26\textwidth}@{}}
\toprule
Family & Structural origin & Sampling and KL & Global behavior \\
\midrule
vMF
& canonical spherical exponential family
& rejection sampler, Bessel normalizer and ratios
& positive at the antipode, intermediate tails and KL \\
Power Spherical
& shifted-cosine power law chosen for tractability
& Beta marginal and reflection, Gamma and digamma KL
& zero at the antipode, narrowest tails and largest KL \\
Spherical Cauchy
& M\"obius image of the uniform law, Student form
& explicit transport, finite or direct-series KL
& positive at the antipode, broadest tails and smallest KL \\
\bottomrule
\end{tabular}
\end{table}

\section{The spherical Cauchy latent layer}
\label{sec:latent}

Let
\[
\SD=\{x\in\R^D:\|x\|=1\},
\qquad
\BD=\{a\in\R^D:\|a\|<1\},
\]
and let $\nuD$ be the uniform probability measure on $\SD$. We use $D$ for the ambient dimension. The sphere has intrinsic dimension $D-1$.

\subsection{Distribution and M\"obius reparameterization}

For $a\in\BD$, define the density with respect to $\nuD$
\begin{equation}
\label{eq:density-ball}
q_a(x)
=
\left(
\frac{1-\|a\|^2}{\|x-a\|^2}
\right)^{D-1},
\qquad x\in\SD.
\end{equation}
The distribution can be parameterized either by the ball parameter $a\in\BD$,
or equivalently by a direction $\mu\in\SD$ and concentration $\rho\in[0,1)$
through $a=\rho\mu$, with $\rho=\|a\|$ and $\mu=a/\|a\|$ whenever $a\neq0$.
In the latter parameterization, 
\begin{equation} \label{eq:density-direction}
q_{\mu,\rho}(x) = \left( \frac{1-\rho^2}{1+\rho^2-2\rho\mu^\top x} \right)^{D-1}.
\end{equation} 
The case $\rho=0$, equivalently $a=0$, is uniform, so the direction $\mu$ is
not identified. The distribution concentrates at $\mu$ as $\rho\to1^-$.

The ball parameter removes the undefined direction at $\rho=0$. An encoder can emit an unconstrained vector $h_\phi(x)\in\R^D$ and map it to
\begin{equation}
\label{eq:ball-encoder}
a_\phi(x)=\frac{h_\phi(x)}{\sqrt{1+\|h_\phi(x)\|^2}}\in\BD.
\end{equation}
The experiments use the equivalent direction-concentration parameterization to remain compatible with the vMF implementation.

For $a\in\BD$, define
\begin{equation}
\label{eq:mobius}
M_a(u)
=
a+(1-\|a\|^2)\frac{u+a}{\|u+a\|^2},
\qquad u\in\SD.
\end{equation}
Kato and McCullagh showed that this M\"obius map pushes $\nuD$ forward to the density $q_a$ in \eqref{eq:density-ball} \PaperCitep{spCauchy}. We use that transport as the VAE reparameterization trick. A pathwise sample is
\[
G\sim\mathcal N(0,I_D),
\qquad
U=\frac{G}{\|G\|},
\qquad
Z=M_a(U).
\]
The map is differentiable in $a$ for every $a\in\BD$. No acceptance decision or implicit gradient is needed. 

\subsection{Matched curvature, angular tails, and KL cost}

The explicit sampler makes spherical Cauchy convenient to use, but its value as a posterior also depends on how it distributes probability. Two posteriors can express the same precision near their shared mode yet retain different amounts of uncertainty elsewhere. We isolate this distinction by matching local curvature, since numerically equal concentration parameters do not have the same meaning across families.

For this comparison, let $\theta=\arccos(\mu^\top x)$ and set $m=D-1$. Dividing each density by its value at the mode removes normalization constants and isolates how quickly probability decays away from the preferred direction. For a vMF concentration $\kappa>0$, choose the spherical Cauchy parameter $\rho$ from
\begin{equation}
\label{eq:kappa-map}
\kappa=\frac{2m\rho}{(1-\rho)^2},
\end{equation}
and choose the Power Spherical exponent $\lambda=2\kappa$.

\begin{proposition}
\label{prop:profile-order}
Under this matching, all three log densities have curvature $-\kappa$ at the mode. Relative to their modal values, they are
\begin{align}
R_{\mathrm{SC}}(\theta)
&=\left[1+\frac{\kappa}{m}(1-\cos\theta)\right]^{-m},\label{eq:sc-profile}\\
R_{\mathrm{vMF}}(\theta)
&=\exp[-\kappa(1-\cos\theta)],\label{eq:vmf-profile}\\
R_{\mathrm{PS}}(\theta)
&=\left(\frac{1+\cos\theta}{2}\right)^{2\kappa}.
\label{eq:ps-profile}
\end{align}
For every $0<\theta<\pi$,
\begin{equation}
\label{eq:profile-order}
R_{\mathrm{PS}}(\theta)
<R_{\mathrm{vMF}}(\theta)
<R_{\mathrm{SC}}(\theta).
\end{equation}
At the antipode this becomes
\begin{equation}
\label{eq:antipodal-order}
R_{\mathrm{PS}}(\pi)=0,
\qquad
R_{\mathrm{vMF}}(\pi)=e^{-2\kappa},
\qquad
R_{\mathrm{SC}}(\pi)=\left(1+\frac{2\kappa}{m}\right)^{-m}>e^{-2\kappa}.
\end{equation}
For draws \(X_{\mathrm{PS}},X_{\mathrm{vMF}},X_{\mathrm{SC}}\) from the three matched-curvature distributions centered at \(\mu\), let
$$
\Theta_j=\arccos(\mu^\top X_j),\qquad
j\in\{\mathrm{PS},\mathrm{vMF},\mathrm{SC}\},
$$
denote their geodesic distances from the mode. Then
\begin{equation}
\label{eq:lr-order}
\Theta_{\mathrm{PS}}
\le_{\mathrm{lr}}
\Theta_{\mathrm{vMF}}
\le_{\mathrm{lr}}
\Theta_{\mathrm{SC}},
\end{equation}
where $\le_{\mathrm{lr}}$ denotes the likelihood-ratio order \PaperCitep{shaked_stochastic_2007}, so that $X\le_{\mathrm{lr}}Y$ if and only if $f_Y(t)/f_X(t)$ is nondecreasing in $t$.
Consequently, for $0<r<\pi$,
\begin{equation}
\label{eq:tail-order}
\Pr(\Theta_{\mathrm{PS}}>r)
<\Pr(\Theta_{\mathrm{vMF}}>r)
<\Pr(\Theta_{\mathrm{SC}}>r),
\end{equation}
and
\begin{equation}
\label{eq:matched-kl-order}
\KL\!\left(\PS_D(\mu,2\kappa)\,\middle\|\,\nuD\right)
>
\KL\!\left(\vMF_D(\mu,\kappa)\,\middle\|\,\nuD\right)
>
\KL\!\left(\SC_D(\mu,\rho)\,\middle\|\,\nuD\right).
\end{equation}
The angular ordering also holds for expectations of every nondecreasing function of the angle.
\end{proposition}

The profile inequalities follow from the elementary bounds $\log(1+y)<y$ and $\log(1-y)<-y$. The stronger distributional conclusion comes from the fact that both adjacent angular likelihood ratios increase with distance from the mode. The KL ordering then follows from a change-of-reference identity. A more concentrated law places greater weight where the broader law has larger density relative to the uniform angular measure, and it also pays a positive KL for the remaining change of law. The full proof is in the \SupportingMaterialName.

The antipodal distinction is stronger than the tail plot alone suggests. For a Power Spherical exponent $\lambda>0$,
\[
q_{\mathrm{PS}}(\theta)\propto\left(\frac{1+\cos\theta}{2}\right)^\lambda,
\qquad
\frac{d}{d\theta}\log q_{\mathrm{PS}}(\theta)=-\lambda\tan(\theta/2).
\]
The density therefore reaches zero at $\theta=\pi$, while its angular score becomes singular as the antipode is approached. More precisely, its density decays like $(\pi-\theta)^{2\lambda}$ near that point. This behavior is imposed by the family for every positive concentration. vMF and spherical Cauchy remain strictly positive at every angle for finite concentration. The Power Spherical profile is well matched to applications where the opposite direction should be excluded a priori. Spherical Cauchy is better matched to applications where remote and antipodal directions should remain possible, although less likely than the modal direction. This matches the behavior of classical directional families that arise naturally from probabilistic or geometric constructions, including vMF, Watson, spherical Cauchy, and Poisson kernel-based distributions.

\begin{figure}[htbp]
    \centering
    \includegraphics[width=0.96\textwidth]{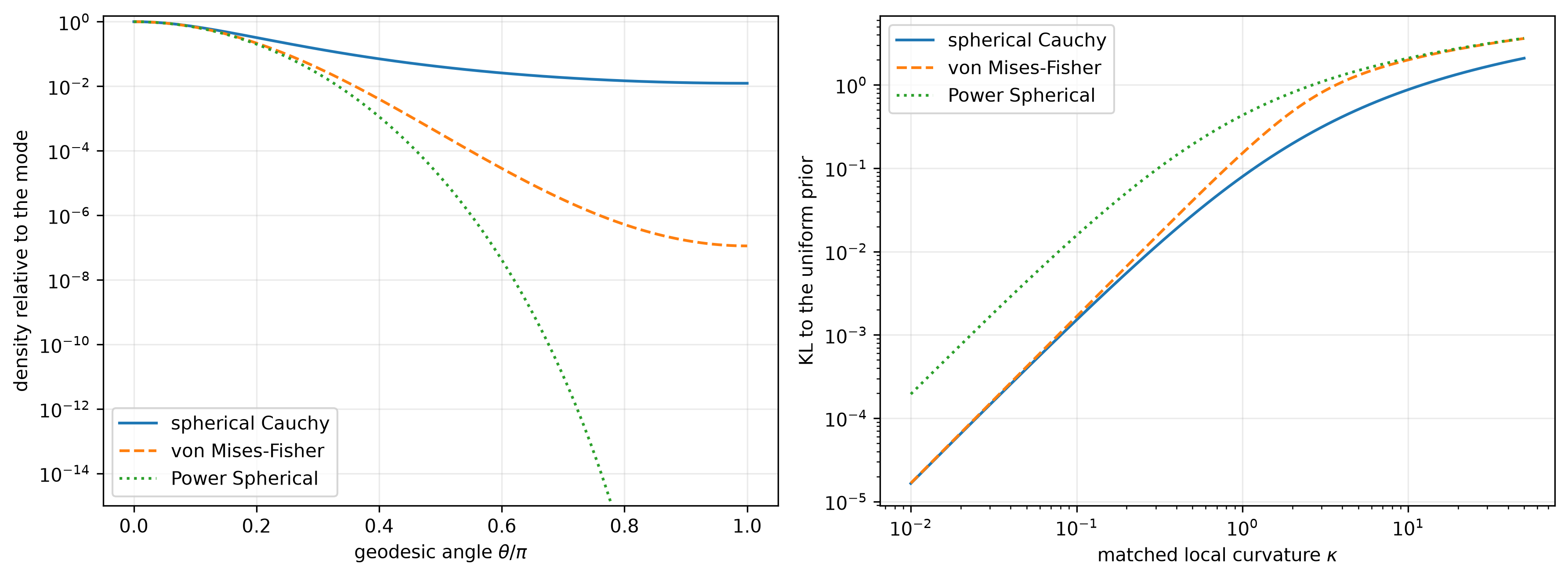}
    \caption{Matched-curvature comparison. The left panel shows density relative to the mode for $D=3$ and $\kappa=8$. Here $\rho=0.5$ and the Power Spherical exponent is $16$. The right panel shows KL divergence to the uniform prior over a range of matched curvature values. Proposition~\ref{prop:profile-order} proves the angular and KL orderings for every $D\ge2$.}
    \label{fig:matched-profiles}
\end{figure}

At the same local curvature, spherical Cauchy has the broadest angular tails and the smallest KL penalty. This reflects the same principle that motivates weakly informative Bayesian modeling: the posterior should express the precision supported near its mode without imposing unnecessary certainty elsewhere. Relative to the common uniform prior, spherical Cauchy therefore carries less global information than the more sharply concentrated alternatives while preserving the same local directional precision and continued access to distant latent directions.

\subsection{VAE objective}

For an observation $x$, let $q_\phi(z\mid x)=\SC_D(a_\phi(x))$, let the prior be $p(z)=\nuD$, and let $p_\theta(x\mid z)$ be a differentiable decoder. We minimize
\begin{equation}
\label{eq:vae-objective}
\mathcal L_{\beta}(x)
=
\E_{q_\phi(z\mid x)}[-\log p_\theta(x\mid z)]
+\beta K_D(\|a_\phi(x)\|),
\end{equation}
where $K_D$ is the KL penalty defined in \eqref{eq:intro-kl} and $\beta>0$ controls the strength of the KL regularization. Proposition~\ref{prop:pairwise-kl} shows that the same evaluator also applies to learned or conditional spherical Cauchy priors.

The reconstruction term encourages latent samples that preserve the information needed by the decoder, while the KL penalizes departures from the uniform prior. Rotational invariance makes this penalty independent of $\mu$, so the cost is determined by how strongly the posterior concentrates around its chosen direction. The encoder learns this balance through the differentiable sampler and the concentration derivative of $K_D$, using a standard decoder and reconstruction loss. It remains to evaluate $K_D$ and its derivative efficiently.

\section{Exact KL theory and numerical evaluation}
\label{sec:exact-kl}

The cost of this evaluation matters because the regularizer and its derivative are needed for every observation at every training step. A special-function ratio, fixed quadrature budget, or regime switch therefore becomes part of the optimization loop. For spherical Cauchy, we can use the sampling map to convert the KL into a one-dimensional angular expectation. The derivation starts with the identity obtained directly from \eqref{eq:mobius},
\[
\|M_a(U)-a\|^2
=\frac{(1-\|a\|^2)^2}{\|U+a\|^2}.
\]
Substitution in \eqref{eq:density-ball} gives
\begin{equation}
\label{eq:transport-density}
q_a(M_a(U))
=\left(\frac{\|U+a\|^2}{1-\|a\|^2}\right)^{D-1}.
\end{equation}
Therefore
\begin{equation}
\label{eq:transport-kl}
K_D(\rho)
=\KL\!\left(\SC_D(\mu,\rho)\,\middle\|\,\nuD\right)=(D-1)\left[-\log(1-\rho^2)+\E_{U\sim\nuD}\log\|U+a\|^2\right].
\end{equation}

\subsection{Direct series and finite even-dimensional formulas}

\begin{theorem}
\label{thm:direct-kl}
For $D\ge2$ and $0\le\rho<1$,
\begin{equation}
\label{eq:direct-kl}
K_D(\rho)
=(D-1)\left[
-\log(1-\rho^2)
-\sum_{j=1}^{\infty}
\frac{(1-D/2)_j}{(D/2)_j}\frac{\rho^{2j}}{j}
\right].
\end{equation}
If the series does not terminate, its $j$th correction term is $O(j^{-D}\rho^{2j})$. The correction is therefore absolutely convergent for every $\rho\le1$.
\end{theorem}

The mechanism is simple. Rotate $a$ to $\rho e_1$ and write $U_1=\cos\Theta$. The logarithm in \eqref{eq:transport-kl} has a Fourier expansion in $\cos(n\Theta)$. Odd moments vanish, while the uniform spherical angle satisfies
\begin{equation}
\label{eq:angular-moment}
\E[\cos(2j\Theta)]
=\frac{(1-D/2)_j}{(D/2)_j}.
\end{equation}
Substitution yields \eqref{eq:direct-kl}. The same identity explains both the polynomial decay in odd dimensions and exact termination in even dimensions. The \SupportingMaterialName{} gives the full beta-integral and interchange arguments.

The theorem separates the singular part of the regularizer from a bounded correction. The only divergence as $\rho\to1^-$ is $-\log(1-\rho^2)$. The correction is controlled by spherical Fourier moments. In even dimensions these moments vanish after a finite order. In odd dimensions they decay polynomially even at the concentration boundary.

The signed correction in \eqref{eq:direct-kl} is the convenient form for exact evaluation. A second rearrangement is more informative for optimization. Expanding the logarithm and collecting equal powers removes all sign changes. The KL becomes a power series with strictly positive coefficients.

\begin{corollary}
\label{cor:positive-expansion}
Let
\[
c_{D,j}=\frac{(1-D/2)_j}{(D/2)_j}.
\]
Then
\begin{equation}
\label{eq:positive-kl-series}
K_D(\rho)
=(D-1)\sum_{j=1}^{\infty}
(1-c_{D,j})\frac{\rho^{2j}}{j},
\end{equation}
and every coefficient is strictly positive. In particular,
\begin{align}
K_D'(\rho)
&=2(D-1)\sum_{j=1}^{\infty}(1-c_{D,j})\rho^{2j-1}>0,\label{eq:positive-first-derivative}\\
K_D''(\rho)
&=2(D-1)\sum_{j=1}^{\infty}(2j-1)(1-c_{D,j})\rho^{2j-2}>0.
\label{eq:positive-second-derivative}
\end{align}
Hence $K_D$ is strictly increasing and strictly convex in $\rho$ on $[0,1)$. It is also strictly convex as a function of $\rho^2$, and
\begin{equation}
\label{eq:global-quadratic-lower}
K_D(\rho)\ge \frac{2(D-1)^2}{D}\rho^2.
\end{equation}
\end{corollary}

The positivity is not an asymptotic statement. The product representation of $c_{D,j}$ gives $|c_{D,j}|<1$ whenever the coefficient is nonzero. Hence every term in \eqref{eq:positive-kl-series} is positive over the full concentration range. The proof and the termwise differentiation argument are given in the \SupportingMaterialName.

Thus concentration has a predictable regularization effect. Increasing $\rho$ always increases the KL, and the marginal cost also increases. The mapping from concentration to KL is therefore one-to-one for every fixed dimension. The global bound \eqref{eq:global-quadratic-lower} also shows that the familiar quadratic behavior near the uniform law remains a valid lower bound over the full parameter range.

\begin{corollary}
\label{cor:even-kl}
Let $D=2q$. Then
\begin{equation}
\label{eq:finite-even-kl}
K_{2q}(\rho)
=(2q-1)\left[
-\log(1-\rho^2)
-\sum_{j=1}^{q-1}
\frac{(1-q)_j}{(q)_j}\frac{\rho^{2j}}{j}
\right].
\end{equation}
The KL is an exact polynomial in $\rho^2$ plus one logarithm. Its concentration gradient is obtained by differentiating the same finite expression and has the same $q-1$ correction terms.
\end{corollary}

The sum terminates because $(1-q)_j=0$ for $j\ge q$. The angular surface weight is then a finite cosine polynomial, so the corresponding Fourier moments vanish above order $q-1$. For $D=128$, the formula contains $63$ nonzero correction terms. Powers of two and many other common latent dimensions are therefore exact finite cases.

The coefficients need no gamma or Pochhammer evaluation. Define
\[
t_j=\frac{(1-D/2)_j}{(D/2)_j}\frac{\rho^{2j}}{j}.
\]
Then
\begin{equation}
\label{eq:term-recurrence}
t_1=\frac{1-D/2}{D/2}\rho^2,
\qquad
t_{j+1}
=t_j\rho^2
\frac{j+1-D/2}{j+D/2}
\frac{j}{j+1}.
\end{equation}
For even $D$, this recurrence reaches zero at the exact termination index. Coefficients can also be precomputed for a fixed dimension and evaluated as a polynomial. A scalar implementation uses $O(D)$ arithmetic and $O(1)$ auxiliary memory. In odd dimensions the cost is $O(N)$, where $N$ is selected by the requested error tolerance.

Termwise differentiation gives
\begin{equation}
\label{eq:kl-gradient}
K_D'(\rho)
=2(D-1)\left[
\frac{\rho}{1-\rho^2}
-\sum_{j=1}^{\infty}
\frac{(1-D/2)_j}{(D/2)_j}\rho^{2j-1}
\right].
\end{equation}
The gradient terminates whenever the KL does. The same derivative gives the ball-parameter gradient by the chain rule and extends continuously to zero.

\subsection{Pairwise KL and hyperbolic parameter geometry}

Let $P_a$ denote the spherical Cauchy law with ball parameter $a\in\BD$. For $a,b\in\BD$, define the pseudohyperbolic distance
\begin{equation}
\label{eq:pseudohyperbolic}
\delta(a,b)
=\left(
\frac{\|a-b\|^2}
{1-2a^\top b+\|a\|^2\|b\|^2}
\right)^{1/2}.
\end{equation}

\begin{proposition}
\label{prop:pairwise-kl}
For every $a,b\in\BD$,
\begin{equation}
\label{eq:pairwise-kl}
\KL(P_a\|P_b)
=K_D\!\left(\delta(a,b)\right)
=\KL(P_b\|P_a).
\end{equation}
The hyperbolic distance in the Poincar\'e-ball model is  \PaperCitep{ganea2018hyperbolic}
\begin{equation}
\label{eq:hyperbolic-distance}
d_{\mathbb H}(a,b)=2\operatorname{atanh}\delta(a,b).
\end{equation}
Hence pairwise KL is a strictly increasing radial function of hyperbolic distance whenever $a\ne b$.
\end{proposition}

The symmetry in \eqref{eq:pairwise-kl} is special to this family. A M\"obius automorphism sends the reference parameter $b$ to the origin and the other parameter to radius $\delta(a,b)$. Invariance of KL under the same invertible transformation then reduces the pairwise problem to the uniform-prior quantity already evaluated by $K_D$. Thus the finite even-dimensional formula and certified odd-dimensional recurrence apply unchanged to learned, conditional, or hierarchical spherical Cauchy priors. The proof is in the \SupportingMaterialName. The square root of the KL divergence can also be shown to define a metric on the spherical Cauchy family (it is an increasing, subadditive transformation of the hyperbolic distance), paralleling the corresponding property of the ordinary Cauchy family.

\subsection{Certified value and gradient evaluation in odd dimensions}

For $D=2q+1$, the coefficients have a fixed sign once $j\ge q$. This yields boundary-stable remainder bounds for both quantities needed during training.

\begin{corollary}
\label{cor:odd-tail}
Let
\[
c_j=\frac{(1-D/2)_j}{(D/2)_j},
\qquad
t_j=c_j\frac{\rho^{2j}}{j},
\qquad
g_j=c_j\rho^{2j-1}.
\]
Truncate both series after $N\ge q-1$ terms and set $n=N+1$. Then
\begin{align}
\left|\sum_{j=n}^{\infty}t_j\right|
&\le
|t_n|\min\left\{
\frac{1}{1-\rho^2},
1+\frac{n}{2q}
\right\},
\label{eq:odd-tail}\\
\left|\sum_{j=n}^{\infty}g_j\right|
&\le
|g_n|\min\left\{
\frac{1}{1-\rho^2},
1+\frac{n}{2q-1}
\right\}.
\label{eq:odd-gradient-tail}
\end{align}
Multiplication by $D-1$ gives the absolute KL error bound. Multiplication by $2(D-1)$ gives the absolute error bound for $K_D'(\rho)$.
\end{corollary}

The geometric parts of the bounds are sharpest away from the concentration boundary. The second parts use polynomial coefficient decay and remain finite as $\rho\to1^-$. Exact value and gradient evaluation are therefore available in every odd dimension without a high-concentration switch. In practice, the coefficients are generated by recurrence and accumulated until the required scaled bound falls below the requested tolerance. Even dimensions stop at the algebraic termination index. The implementation uses \texttt{log1p} for the logarithmic term and can cache coefficients for every fixed dimension.

\subsection{Integral representation and structural consequences}

Define
\begin{equation}
\label{eq:z-def}
z(\rho)=\frac{4\rho}{(1+\rho)^2}
\end{equation}
and
\begin{equation}
\label{eq:J-def}
J_D(z)
=\int_0^1
\frac{t^{D-2}}{1-t}
\left[1-\left(\frac{1-z}{1-zt}\right)^{(D-1)/2}\right]dt.
\end{equation}

\begin{proposition}
\label{prop:integral}
For $D\ge2$ and $0\le\rho<1$,
\begin{equation}
\label{eq:integral-kl}
K_D(\rho)
=(D-1)\left[
\log\left(\frac{1-\rho}{1+\rho}\right)
+J_D(z(\rho))
\right].
\end{equation}
The integral is finite for every $0\le\rho<1$.
\end{proposition}

The integral is derived from a logarithmic derivative of a Gauss hypergeometric function. It is not the preferred exact evaluator, but it is useful for asymptotic analysis, independent checking, and approximation design. In fact, for every integer $D\ge2$, the integral $J_D$ can be evaluated in elementary functions, although the resulting expressions become complicated rather quickly. For our purposes, it is sufficient to exploit the compact closed forms available up to $D=6$, which are recorded in the \SupportingMaterialName.

\begin{proposition}
\label{prop:asymptotics}
For the spherical Cauchy KL,
\begin{align}
K_D(\rho)
&=\frac{2(D-1)^2}{D}\rho^2+O(\rho^4),
&&\rho\to0,\label{eq:small-rho}\\
K_D(\rho)
&=(D-1)\left[
-\log(1-\rho^2)+2\log 2
+\psi\!\left(\frac{D-1}{2}\right)-\psi(D-1)
\right]+o(1),
&&\rho\to1^-,\label{eq:high-rho}\\
K_D(\rho)
&=(D-1)\log\left(\frac{1+\rho^2}{1-\rho^2}\right)+o(D),
&&D\to\infty,
\label{eq:large-D}
\end{align}
where the first two limits hold at fixed $D$ and the last holds at fixed $0<\rho<1$.
\end{proposition}

The first expansion also gives
$\nabla_a K_D(\|a\|)=4(D-1)^2a/D+O(\|a\|^3)$ near $a=0$.

The large-dimensional limit suggests that a fixed concentration should become more expensive as the sphere grows. The next result shows that this is not only asymptotic. Both the total KL and the KL per intrinsic dimension increase at every finite dimension.

\begin{proposition}
\label{prop:dimension-monotonicity}
For every fixed $0<\rho<1$ and $D\ge2$,
\begin{equation}
\label{eq:normalized-d-monotonicity}
\frac{K_{D+1}(\rho)}{D}
>
\frac{K_D(\rho)}{D-1}
>0.
\end{equation}
Consequently,
\begin{equation}
\label{eq:d-monotonicity}
K_{D+1}(\rho)>K_D(\rho).
\end{equation}
\end{proposition}

The same numerical value of $\rho$ therefore represents a more informative posterior in a higher ambient dimension. Concentration values should not be compared across dimensions without also considering the KL budget. In a family of models with similar effective regularization, the learned $\rho$ will typically decrease as $D$ grows. The proof uses the compact integral and appears in the \SupportingMaterialName.

\subsection{A finite approximation for odd dimensions}

Monotonicity in ambient dimension makes the terminating even-dimensional formulas useful beyond their exact parity class. If $D$ is odd, both adjacent dimensions are even. We therefore define the even-neighbor approximation
\begin{equation}
\label{eq:neighbor-average}
K_D^{\mathrm{nbr}}(\rho)
=\frac{K_{D-1}(\rho)+K_{D+1}(\rho)}{2}.
\end{equation}

\begin{corollary}
\label{cor:neighbor-average}
For odd $D\ge3$ and $0\le\rho<1$, the two values in \eqref{eq:neighbor-average} are exact finite formulas. Moreover,
\begin{equation}
\label{eq:neighbor-certificate}
\left|K_D^{\mathrm{nbr}}(\rho)-K_D(\rho)\right|
\le
\frac{K_{D+1}(\rho)-K_{D-1}(\rho)}{2}.
\end{equation}
Let $c_{d,j}=(1-d/2)_j/(d/2)_j$, with $c_{d,j}=0$ after the even-dimensional termination index. Then
\begin{equation}
\label{eq:neighbor-polynomial}
K_D^{\mathrm{nbr}}(\rho)
=(D-1)\left[
-\log(1-\rho^2)
-\sum_{j=1}^{(D-1)/2}
\overline c_{D,j}\frac{\rho^{2j}}{j}
\right],
\end{equation}
where
\begin{equation}
\label{eq:neighbor-coefficients}
\overline c_{D,j}
=\frac{(D-2)c_{D-1,j}+Dc_{D+1,j}}{2(D-1)}.
\end{equation}
Thus the average can be evaluated as one precomputed finite polynomial and one logarithm.
\end{corollary}

For $D=3,5$ we retain the compact elementary exact formulas, so the even-neighbor approximation is needed only for odd $D\ge7$. Numerical maximization over $\rho\in[0,1)$ for odd $D\in\{7,9,\ldots,199\}$ gives a largest observed absolute error of $0.00109950$. It occurs at $D=7$ as $\rho\to1^-$. The error decreases approximately cubically with dimension over the tested range. This route is finite and highly accurate, although its arithmetic cost still grows linearly with $D$.

We use the following finite rule when an approximation is acceptable:
\begin{equation}
\label{eq:finite-rule}
K_D^{\mathrm{finite}}(\rho)=
\begin{cases}
K_D(\rho)\text{ from \eqref{eq:finite-even-kl}}, & D\text{ even},\\
K_D(\rho)\text{ from exact elementary formulas}, & D\in\{3,5\},\\
K_D^{\mathrm{nbr}}(\rho), & D\ge7\text{ odd}.
\end{cases}
\end{equation}

\subsection{Analytic envelopes and a constant-cost approximation}

The finite rule has predetermined cost for each dimension, but its polynomial degree grows with $D$. A dimension-independent route remains useful for large $D$ and for direct runtime comparisons with other closed-form posterior families. The compact integral gives two analytic envelopes whose gap is independent of concentration.

Set
\begin{equation}
\label{eq:wD}
w_D
=\psi(D-1)-\psi\!\left(\frac{D-1}{2}\right)-\log 2
=\frac{1}{2(D-1)}+O(D^{-2}).
\end{equation}

\begin{proposition}
\label{prop:bracket}
For $D\ge2$ and $0\le\rho<1$,
\begin{align}
(D-1)\left[
\log\left(\frac{2(1+\rho^2)}{1-\rho^2}\right)
+\psi\!\left(\frac{D-1}{2}\right)-\psi(D-1)
\right]
&\le K_D(\rho),\label{eq:lower-bound}\\
K_D(\rho)
&\le(D-1)\log\left(\frac{1+\rho^2}{1-\rho^2}\right).
\label{eq:upper-bound}
\end{align}
The bracket width is $(D-1)w_D=1/2+O(D^{-1})$.
\end{proposition}

The Laplace-motivated weight
\begin{equation}
\label{eq:laplace-weight}
\omega(\rho)=\left(\frac{2\rho}{1+\rho^2}\right)^2
=\left(\frac{z}{2-z}\right)^2
\end{equation}
interpolates between the two envelopes. It gives
\begin{equation}
\label{eq:surrogate}
\widehat K_D(\rho)
=(D-1)\left[
\log\left(\frac{1+\rho^2}{1-\rho^2}\right)
-w_D\omega(\rho)
\right].
\end{equation}
The surrogate lies inside the bracket, is exact at $\rho=0$, and has the exact leading divergence and constant offset as $\rho\to1^-$. For fixed $\rho<1$,
\begin{equation}
\label{eq:surrogate-asymptotic}
K_D(\rho)-\widehat K_D(\rho)=O(D^{-1}).
\end{equation}
The \SupportingMaterialName{} shows that the same weight matches the first correction in the large-dimensional expansion. Its value does not depend on a series length or quadrature budget.

Table~\ref{tab:evaluator-map} summarizes the resulting evaluator map. The even-neighbor rule is the preferred finite approximation in odd dimensions. The Laplace-weighted route remains useful when constant cost is more important than its larger approximation error.

\begin{table}[htbp]
\centering
\small
\caption{Recommended KL evaluation routes. Exact to tolerance means that the requested absolute error is certified by Corollary~\ref{cor:odd-tail}.}
\label{tab:evaluator-map}
\begin{tabular}{lll}
\toprule
Ambient dimension & Route & Cost and guarantee \\
\midrule
$D=2q$ & finite recurrence & $q-1$ terms, exact \\
$D\in\{3,5\}$ & elementary formula & constant cost, exact \\
odd $D\ge7$ & certified recurrence & $N$ terms, exact to tolerance \\
odd $D\ge7$ & even-neighbor average & finite $O(D)$ approximation \\
any $D$ & Laplace-weighted surrogate & constant-cost approximation \\
\bottomrule
\end{tabular}
\end{table}

Independent $80$-digit calculations agree with both the compact integral and direct spherical-angle integration to below $5\times10^{-67}$ on the recorded grid. Float32 and float64 values and gradients were checked through dimension $2048$. The largest observed even-neighbor error over odd dimensions through $199$ is $1.10\times10^{-3}$, compared with $3.56\times10^{-2}$ for the constant-cost Laplace route. Full grids and figures are in the \SupportingMaterialName.

\section{Experiments}
\label{sec:experiments}

The experiments test numerical stability, computational cost, controlled remote mass, and end-to-end reconstruction. Code, configurations, seed-level results, and manifests are available in the \href{https://github.com/lsablica/SC-VAE}{SC-VAE repository}. Full protocols and secondary diagnostics are in the \SupportingMaterialName.

\subsection{Latent-layer stress test}
\label{subsec:latent-benchmark}

We time a complete forward and backward latent-layer step for the direct spherical Cauchy evaluator \eqref{eq:finite-even-kl}, the finite even-neighbor \eqref{eq:neighbor-polynomial} and constant-cost Laplace routes \eqref{eq:surrogate}, robust and original vMF implementations, and the official Power Spherical path. The robust vMF implementation replaces the original special-function evaluations by stable continued-fraction and log-space calculations following \PaperCitep{hornik2014maximum,movMF}. Concentrations are matched by modal curvature with $\kappa=10$, and the dimensions are
\[
D\in\{8,16,32,64,128,256,512,1024,2048\}.
\]
The even-neighbor method uses $D+1$ instead of $D$.
The \supportingMaterialName{} gives the warm-up, synchronization, batch sizes, backend details, and complete failure criteria.

\begin{figure}[htbp]
    \centering
    \includegraphics[width=\textwidth]{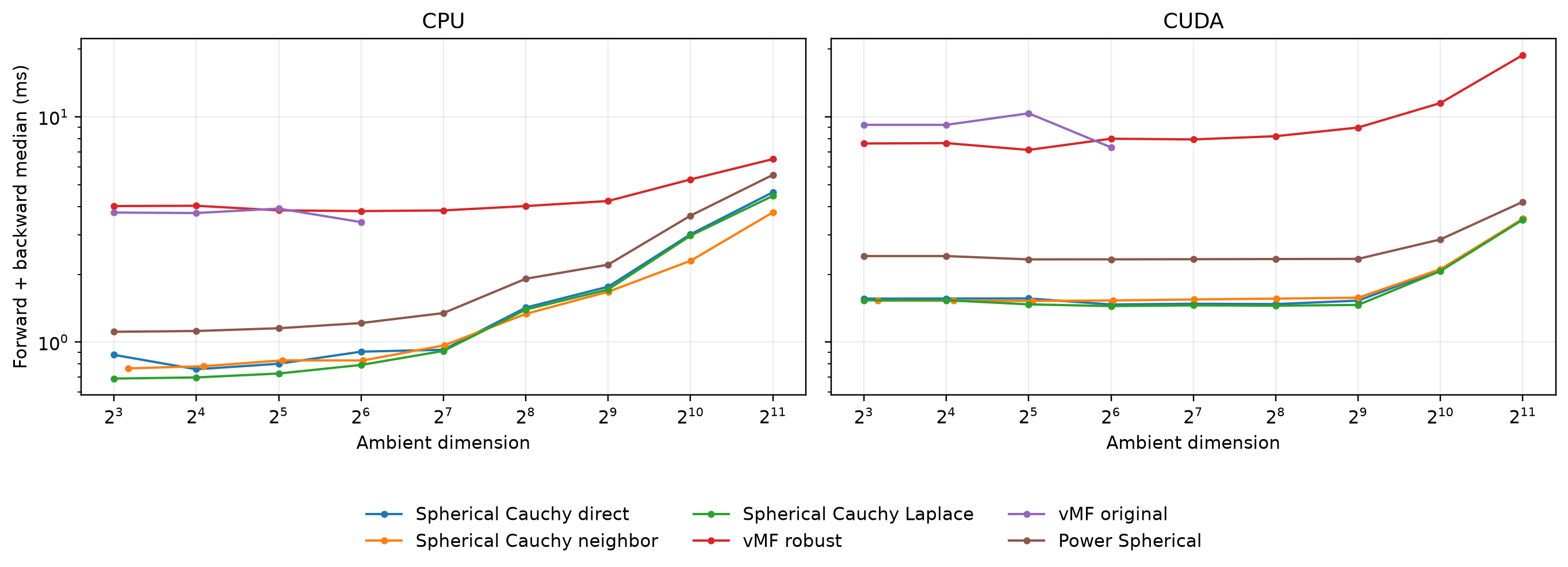}
    \caption{Median time per latent-layer gradient step. Spherical Cauchy direct, even-neighbor, and Laplace routes are compared with original and robust vMF and Power Spherical implementations.}
    \label{fig:latent-runtime}
\end{figure}

Every spherical Cauchy route, robust vMF, and Power Spherical succeeds in all nine dimensions. The original vMF route becomes nonfinite from $D=128$. At $D=128$, direct spherical Cauchy takes $0.925$ ms on CPU and $1.481$ ms on CUDA, compared with $1.347$ and $2.335$ ms for Power Spherical and $3.846$ and $7.945$ ms for robust vMF. Thus the direct route is $1.46$ times faster than Power Spherical and $4.16$ times faster than robust vMF on CPU. The corresponding CUDA factors are $1.58$ and $5.36$. Thus, the additional robustness and exactness of spherical Cauchy do not come at a computational cost. On the contrary, the direct evaluator is among the fastest of the compared posterior routes in the recorded benchmark, on both CPU and GPU.

\subsection{Low-dimensional representations on MNIST}
\label{subsec:mnist}

We compare Gaussian, vMF, spherical Cauchy, and Power Spherical latent variables on MNIST \PaperCitep{lecun1998mnist} with the same convolutional backbone. The reported dimension $p\in\{2,3,5,10,20\}$ is the Gaussian latent dimension and the intrinsic spherical dimension, so spherical models use ambient dimension $D=p+1$. We also note that the standard diagonal Gaussian posterior uses $2p$ encoder outputs for its mean and variance, compared with $p+2$ for the direction and concentration of the spherical posteriors, so this difference in posterior parameterization grows with $p$. Every configuration uses the same five seed indices and deterministic data ordering. Held-out reconstruction is the primary cross-family metric because the models have different KL geometries and sensitivities, so their total objectives are not directly comparable. In particular, the Gaussian posterior is regularized against a Gaussian prior from the same Euclidean family, whereas the spherical posteriors are regularized against the uniform spherical prior. The results are summarized in Table~\ref{tab:mnist}. A dagger marks the prespecified paired comparison with the next-lowest method after Benjamini-Hochberg correction across dimensions.

\begin{table}[t]
\centering
\caption{Held-out MNIST reconstruction loss. Values are means and standard deviations over five paired seeds. Lower is better.}
\label{tab:mnist}
\small
\setlength{\tabcolsep}{5.5pt}
\renewcommand{\arraystretch}{1.08}
\begin{tabular}{ccccc}
\toprule
$p$ & Gaussian & Spherical Cauchy & vMF & Power Spherical \\
\midrule
2  & \ms{133.07}{0.59} & \bmsd{130.29}{0.40} & \ms{132.03}{1.32} & \ms{132.53}{0.82} \\
3  & \ms{118.65}{1.07} & \bmsd{116.44}{0.63} & \ms{118.25}{0.37} & \ms{119.04}{0.48} \\
5  & \ms{102.40}{1.25} & \bmsd{100.22}{0.21} & \ms{104.55}{0.44} & \ms{106.09}{0.65} \\
10 & \ms{81.07}{0.23}  & \bmsd{80.30}{0.38}  & \ms{83.38}{0.36} & \ms{84.12}{0.54} \\
20 & \ms{75.08}{0.27}  & \bmsd{74.43}{0.30}  & \ms{97.25}{45.06} & \ms{82.29}{6.02} \\
\bottomrule
\end{tabular}
\end{table}

Spherical Cauchy has the lowest mean reconstruction loss at every dimension. The paired improvement over the next-lowest method ranges from $0.65$ to $2.18$ loss units. Every paired $95\%$ interval excludes zero, and all adjusted values satisfy $q\le0.0233$. The narrowing reconstruction gap to Gaussian at larger $p$ is consistent with its additional flexibility. The complete objective, KL, interval, and training-stability results are in the \supportingMaterialName.

\begin{figure}[htbp]
    \centering
    \includegraphics[width=0.50\textwidth]{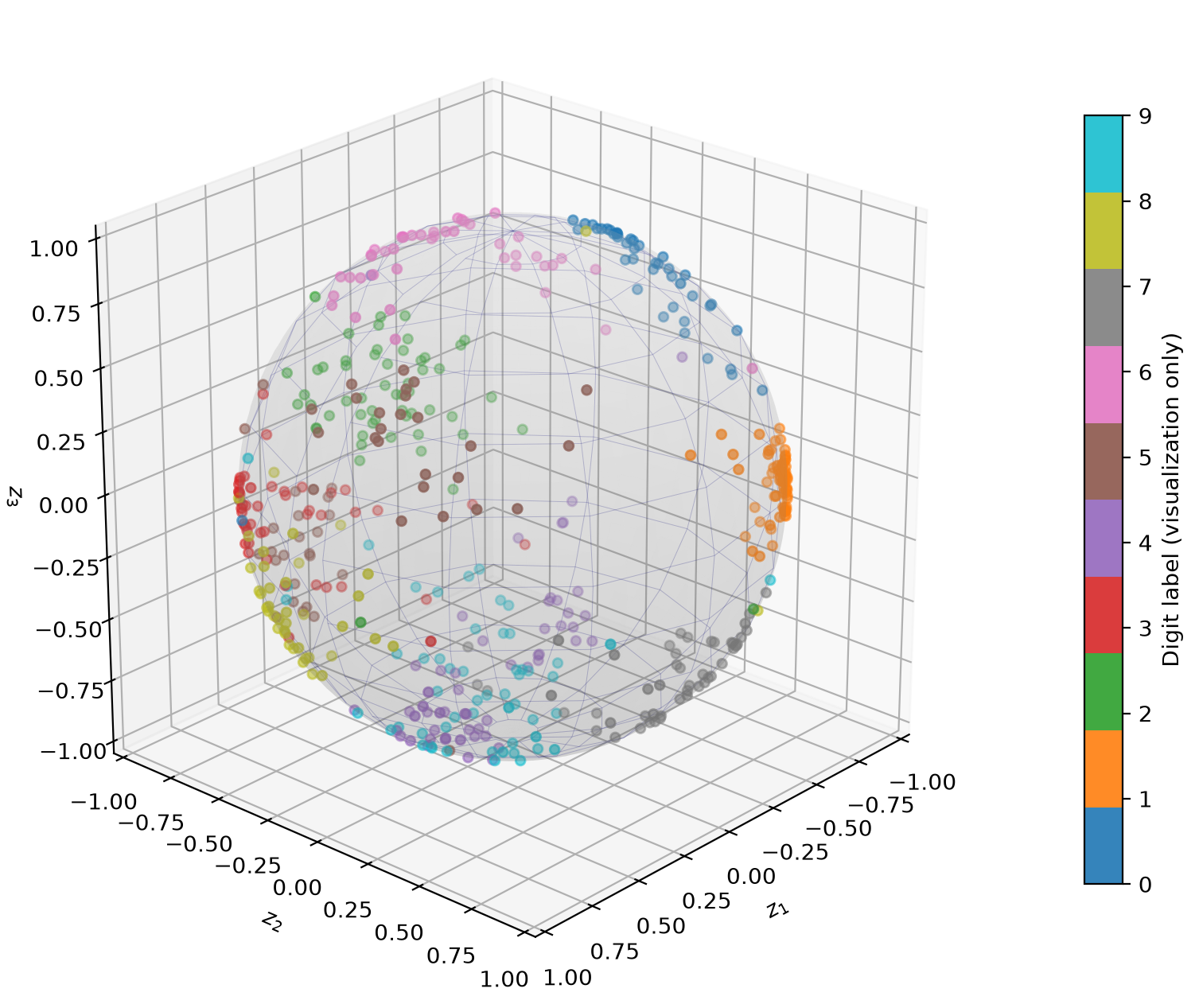}\hfill
    \includegraphics[width=0.47\textwidth]{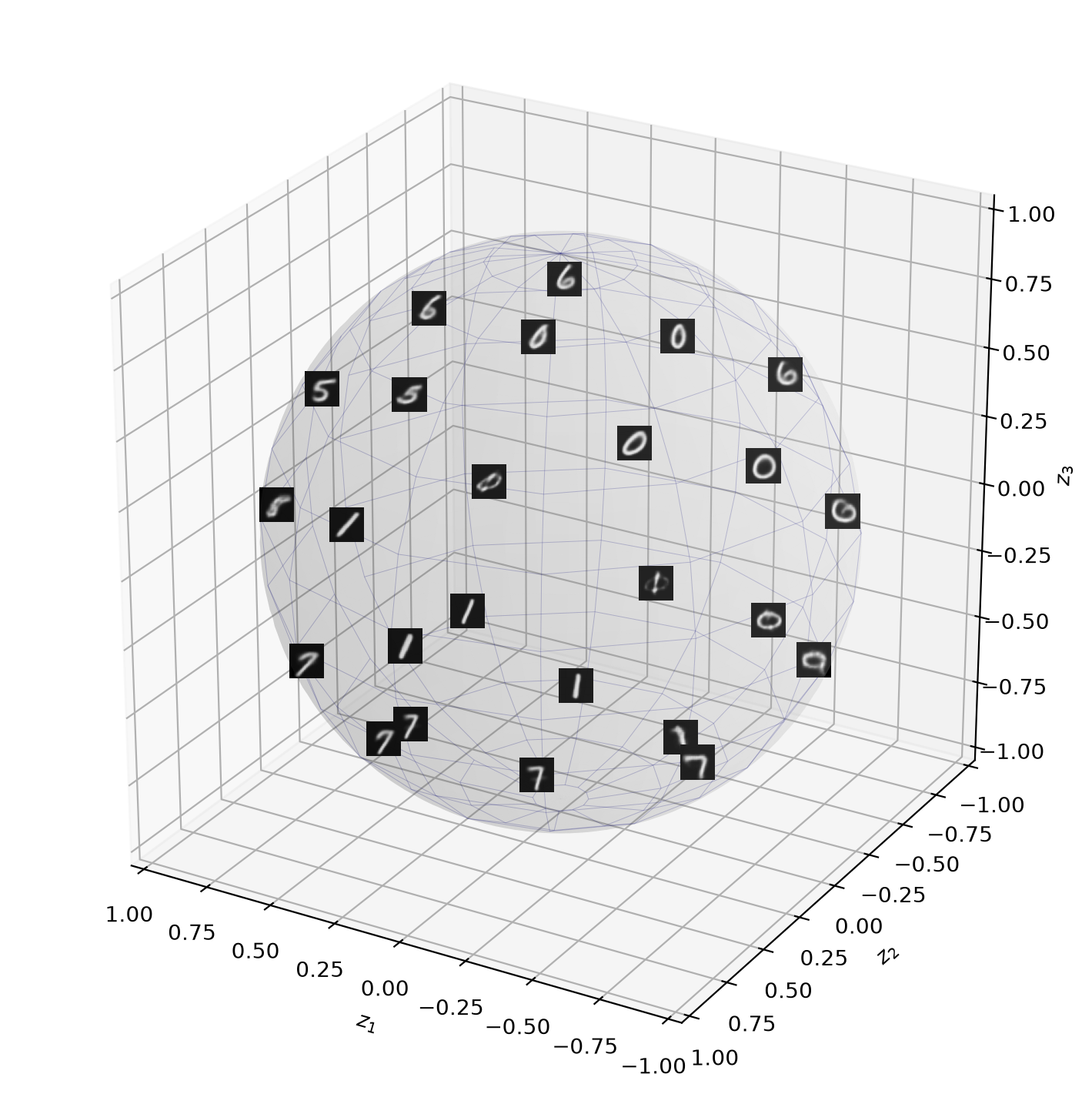}
    \caption{Global organization of a spherical Cauchy VAE on $\mathbb S^2$. The left panel colors posterior samples by digit label for visualization only. The right panel decodes uniformly sampled spherical locations. An interactive version is available on our \href{https://lsablica.github.io/SC-VAE/}{GitHub page}.}
    \label{fig:mnist-organization}
\end{figure}

Figure~\ref{fig:mnist-organization} shows that the unsupervised posterior organizes digit classes into coherent angular regions and that uniform locations decode to recognizable digits over broad parts of the sphere. Reconstructions and a great-circle digit interpolation are shown in the \supportingMaterialName.

\subsection{Controlled directional contamination}
\label{subsec:contamination}

\begin{figure}[htbp]
    \centering
    \includegraphics[width=\textwidth]{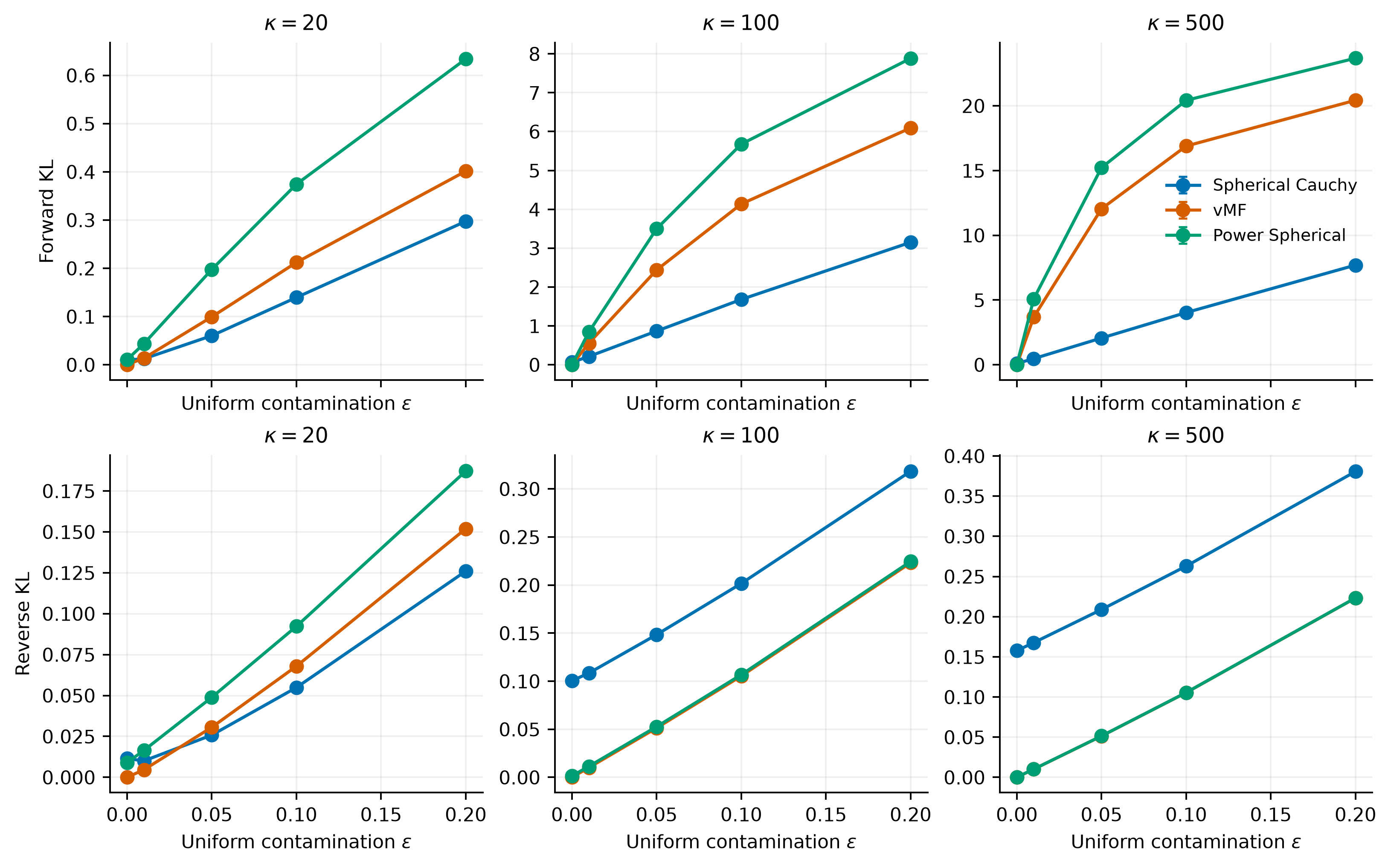}
    \caption{KL of fitted spherical families against uniform contamination. Forward KL in the upper row selects spherical Cauchy for every positive contamination level. The lower row shows the contrasting mode-seeking behavior of reverse KL.}
    \label{fig:contamination-transition}
\end{figure}

Real directional data need not follow a single idealized family exactly and may contain a small amount of diffuse or atypical probability mass. To isolate the effect of such remote mass, we fit spherical Cauchy, vMF, and Power Spherical approximations to
\begin{equation}
\label{eq:contaminated-target}
p_{\epsilon,\kappa}(z)
=(1-\epsilon)\vMF(z\mid\mu,\kappa)+\epsilon\nuD(z)
\end{equation}
on $\mathbb S^{32}$, with $\kappa\in\{20,100,500\}$ and $\epsilon\in\{0,.01,.05,.10,.20\}$. At $\epsilon=0$, vMF is best for every concentration. Under forward KL, spherical Cauchy is best in all twelve positive-contamination cells. At $\kappa=100$ and $\epsilon=.10$, its KL is $1.6811$, compared with $4.1368$ for vMF and $5.6777$ for Power Spherical. Figure~\ref{fig:contamination-transition} shows that the transition occurs already at one percent contamination. Under reverse KL, vMF remains preferred for the sharper targets because the objective is mode seeking and the dominant component of the target is itself vMF, so the small uniform component can be largely ignored. Spherical Cauchy nevertheless becomes best from $\epsilon=.05$ at $\kappa=20$. Complete forward- and reverse-KL results are reported in the
\supportingMaterialName.

Because the uncontaminated target is itself vMF, the crossover is particularly informative. Once even a small diffuse component is introduced, a single spherical Cauchy law can retain the sharp central mode while accommodating remote probability mass substantially better under forward KL than the lighter-tailed alternatives.

\subsection{Viewpoint-gap generalization on smallNORB}
\label{subsec:smallnorb}

In this final experiment, the smallNORB dataset provides a periodic viewpoint factor with known azimuth \PaperCitep{lecun2004norb}. We reserve one official training instance per category for validation and remove the contiguous azimuth sector $320^\circ$, $340^\circ$, $0^\circ$, and $20^\circ$ from training. Thus, the models never see images from this entire viewpoint sector during optimization, and we compare how well they reconstruct test images acquired at these held-out angles. The primary endpoint is reconstruction NLL on the corresponding viewpoints of the official test instances. The intrinsic latent dimension is $32$, so spherical models use $D=33$. An isotropic $32$-dimensional Gaussian has one scalar scale and is parameter matched to the spherical posteriors. All four families share the same $6.35$ million-parameter architecture, optimization schedule, checkpoint rule, and five seed indices.

\begin{figure}[htbp]
    \centering
    \includegraphics[width=\textwidth]{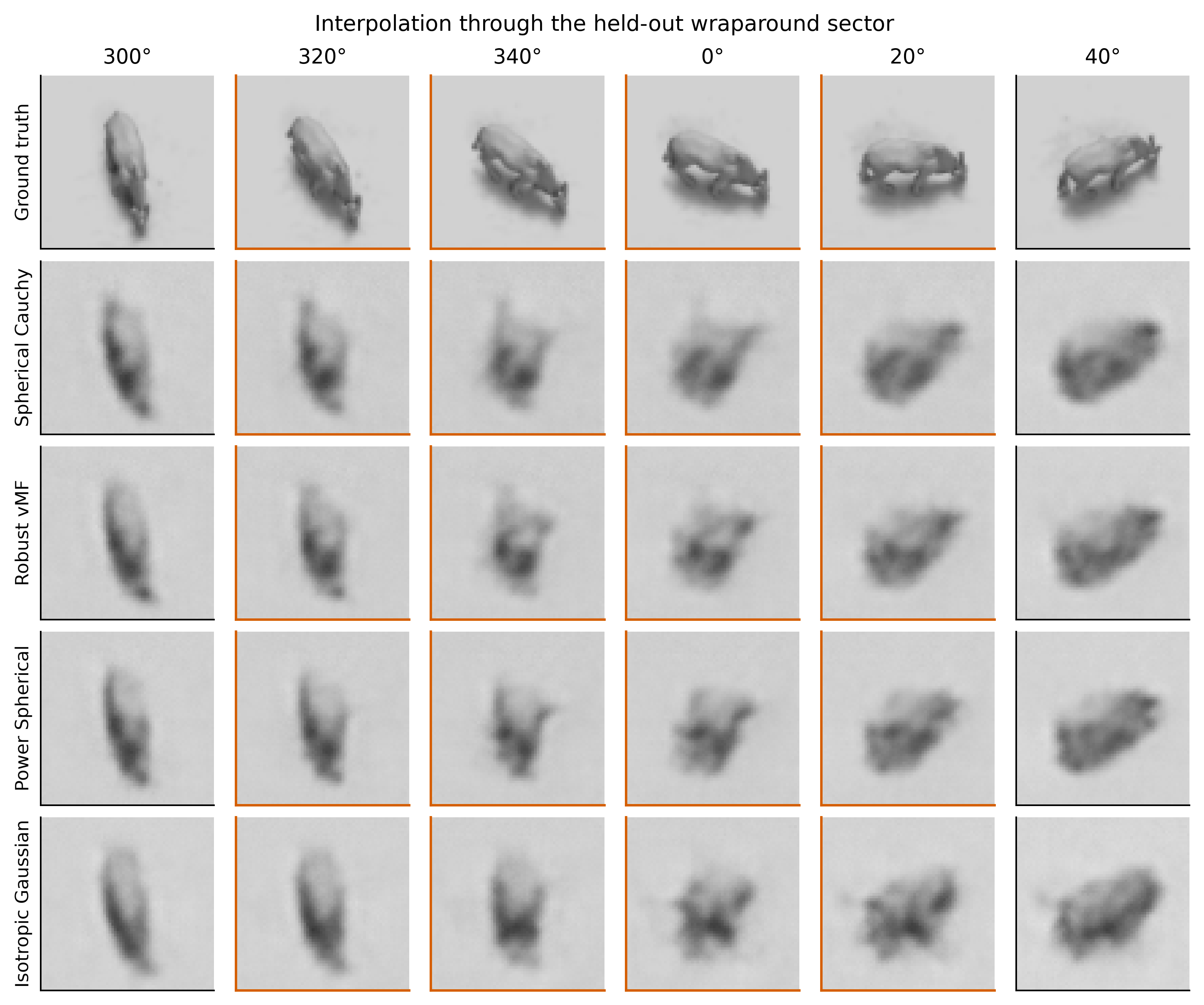}
    \caption{Posterior-location reconstructions through the held-out wraparound sector. Orange borders mark the four azimuths absent from training. The observed viewpoints at $300^\circ$ and $40^\circ$ anchor the path.}
    \label{fig:smallnorb-gap}
\end{figure}

Spherical Cauchy has the lowest gap NLL in every seed and improves on robust vMF by $3.6\%$. The paired difference is $-4.65$, with a $95\%$ interval from $-6.37$ to $-2.93$ and $p=.0017$. It also gives the lowest observed-view NLL, so the held-out-sector gain does not trade off ordinary reconstruction. Figure~\ref{fig:smallnorb-gap} shows a representative wraparound path. Thus, when an entire contiguous viewpoint sector is absent from training, the spherical Cauchy VAE reconstructs previously unseen test objects at those held-out viewpoints better than the same architecture equipped with vMF, Power Spherical, or an isotropic Gaussian posterior. Additional pose, distance, interpolation, factor, and training diagnostics are reported in the \supportingMaterialName. 

\begin{table}[t]
\centering
\small
\caption{smallNORB reconstruction over five paired seeds. Lower is better. The dagger marks the paired primary-endpoint comparison with robust vMF.}
\label{tab:smallnorb}
\begin{tabular}{lrrr}
\toprule
Model & Gap NLL & Gap MSE & Observed NLL \\
\midrule
Gaussian & \ms{173.14}{2.77} & \ms{.00338}{.00005} & \ms{147.32}{2.29} \\
Power Spherical & \ms{130.11}{2.71} & \ms{.00254}{.00005} & \ms{112.44}{2.16} \\
Spherical Cauchy & \bmsd{122.90}{1.05} & \bmsd{.00240}{.00002} & \bmsd{106.18}{.88} \\
Robust vMF & \ms{127.55}{1.10} & \ms{.00249}{.00002} & \ms{110.34}{.48} \\
\bottomrule
\end{tabular}
\end{table}

\section{Discussion}
\label{sec:discussion}

A spherical latent space preserves angular structure in a probabilistic representation, but the choice of posterior determines both the cost of training and how uncertainty is expressed. Spherical Cauchy allows precise local information without strongly excluding remote directions, and our sampler and KL evaluator make this choice practical within the standard VAE objective.

Both computational components come from the same M\"obius transport. It gives an exact reparameterization and a direct KL expansion whose correction is finite in every even ambient dimension and remains certifiably evaluable in odd dimensions. Positive coefficients make the KL increasing and convex in concentration, while dimension monotonicity explains how the information cost changes when the latent sphere grows. The pairwise identity extends the same evaluator to learned or conditional spherical Cauchy priors and identifies KL as a symmetric radial function of hyperbolic parameter distance.

The matched-curvature ordering gives the posterior a clear geometric role. Spherical Cauchy retains the same local precision as vMF and Power Spherical while assigning more probability to distant directions and paying a smaller KL cost to the uniform prior. The controlled contamination study supports this distinction. vMF is recovered when the target is exactly vMF, while even a small remote component changes the best forward-KL approximation to spherical Cauchy.

The numerical results show that exactness does not require a slow latent layer. The fused direct implementation is faster than Power Spherical and robust vMF in the recorded benchmark. Spherical Cauchy also gives the lowest reconstruction loss across all tested MNIST dimensions and the lowest held-out viewpoint-gap NLL on smallNORB.

Mixtures, structured priors, and joint Euclidean-spherical latent spaces are natural extensions. The open-ball parameterization is convenient for these directions because it represents location and concentration with one smooth vector.
\section*{Statements and declarations}

\ifSCVAEJournalVersion\else
\paragraph{Funding}
\PaperFunding
\fi

\paragraph{Data and code availability}
Code, configurations, seed-level summaries and raw benchmark tables are available in the \href{https://github.com/lsablica/SC-VAE}{SC-VAE repository}. MNIST and smallNORB are publicly available, and the repository contains deterministic preprocessing and split definitions.

\bibliographystyle{plainnat}
\bibliography{scvae}

\clearpage
\appendix
\setcounter{theorem}{0}
\section{Proofs and additional KL identities}
\label{app:proofs}

Throughout this appendix, $D\ge2$ is the ambient dimension, $m=D-1$, and $\nuD$ is the uniform probability measure on $\SD$.

\subsection{M\"obius transport identities}

The pushforward result used in the main paper is due to Kato and McCullagh (2020). We record only the identities needed for the KL derivation. For
\[
M_a(u)=a+(1-\|a\|^2)\frac{u+a}{\|u+a\|^2},
\]
direct calculation gives
\[
\|M_a(u)-a\|^2
=\frac{(1-\|a\|^2)^2}{\|u+a\|^2},
\]
so
\[
q_a(M_a(u))
=\left(\frac{\|u+a\|^2}{1-\|a\|^2}\right)^{D-1}.
\]
These formulas recover the pushforward by change of variables and reduce the KL to the uniform expectation used in Theorem~\ref{thm:direct-kl}.

\subsection{Matched-curvature tail and KL ordering}

\begin{proof}[Proof of Proposition~\ref{prop:profile-order}]
Set
\[
s=1-\cos\theta,
\qquad 0\le s\le2.
\]
For spherical Cauchy, the density relative to its value at the mode is
\[
R_{\mathrm{SC}}(\theta)
=
\left(
\frac{(1-\rho)^2}
{(1-\rho)^2+2\rho s}
\right)^m
=
\left(
1+\frac{\kappa}{m}s
\right)^{-m},
\]
where the last equality uses
\[
\kappa=\frac{2m\rho}{(1-\rho)^2}.
\]
The corresponding vMF and Power Spherical profiles are
\[
R_{\mathrm{vMF}}(\theta)
=
e^{-\kappa s},
\qquad
R_{\mathrm{PS}}(\theta)
=
\left(1-\frac{s}{2}\right)^{2\kappa},
\]
where the second identity uses $\lambda=2\kappa$.

Since
\[
s=\frac{\theta^2}{2}+O(\theta^4),
\]
we have
\[
\log R_{\mathrm{SC}}(\theta)
=
-\kappa s+O(s^2),
\qquad
\log R_{\mathrm{vMF}}(\theta)
=
-\kappa s,
\qquad
\log R_{\mathrm{PS}}(\theta)
=
-\kappa s+O(s^2).
\]
Hence all three log profiles satisfy
\[
\log R(\theta)
=
-\frac{\kappa}{2}\theta^2+O(\theta^4),
\]
so their log-density curvature at the mode is $-\kappa$.

For $0<s<2$, the elementary inequalities
\[
\log(1-y)<-y,
\qquad 0<y<1,
\]
and
\[
\log(1+y)<y,
\qquad y>0,
\]
give
\[
2\kappa\log(1-s/2)
<
-\kappa s
<
-m\log(1+\kappa s/m).
\]
Exponentiating proves \eqref{eq:profile-order}.

The induced angular densities all contain the same surface factor
$\sin^{D-2}\theta$, so their likelihood ratios have the same
monotonicity as the corresponding profile ratios. Since $s$ is
strictly increasing in $\theta$ on $(0,\pi)$, direct differentiation
gives
\[
\frac{d}{ds}
\log\frac{R_{\mathrm{vMF}}}{R_{\mathrm{PS}}}
=
\frac{\kappa s}{2-s}>0,
\qquad
\frac{d}{ds}
\log\frac{R_{\mathrm{SC}}}{R_{\mathrm{vMF}}}
=
\frac{\kappa^2s}{m+\kappa s}>0.
\]
This proves \eqref{eq:lr-order}. Likelihood-ratio order implies
stochastic order, which gives \eqref{eq:tail-order} and the
expectation statement.

It remains to compare the KL divergences. Rotational symmetry reduces
the full spherical KL to the KL between the corresponding angular
laws. Consider either adjacent pair in \eqref{eq:lr-order}, let $P$
be the smaller-angle law and $Q$ the larger-angle law, and let $U$
be the angular law induced by $\nuD$. The ratio $q/u$ is a positive
constant times the relative profile of $Q$, and is therefore strictly
decreasing in the angle. Since $P$ is stochastically smaller than $Q$,
\[
\E_P\log(q/u)
\ge
\E_Q\log(q/u).
\]
Consequently,
\[
\begin{aligned}
\KL(P\|U)
&=
\KL(P\|Q)+\E_P\log(q/u) \\
&>
\E_Q\log(q/u)
=
\KL(Q\|U),
\end{aligned}
\]
where the strict inequality uses $P\ne Q$ and hence
$\KL(P\|Q)>0$. Applying this argument to both adjacent pairs proves
\eqref{eq:matched-kl-order}.
\end{proof}

\subsection{Angular Fourier moments}

\begin{lemma}
\label{lem:angular-moments}
Let $U\sim\nuD$ and write $U_1=\cos\Theta$, with $\Theta\in[0,\pi]$. Then
\begin{equation}
\label{eq:supp-angular-moment}
\E[\cos(2j\Theta)]
=\frac{(1-D/2)_j}{(D/2)_j},
\qquad j=0,1,\ldots
\end{equation}
All odd cosine moments vanish.
\end{lemma}

\begin{proof}
Let $U\sim\nuD$ and write $U_1=\cos\Theta$. Using the standard
Gaussian representation $U=G/\|G\|$, with
$G\sim\mathcal N(0,I_D)$, we have
\[
U_1^2
=
\frac{G_1^2}{G_1^2+\cdots+G_D^2}
\sim
\operatorname{Beta}\left(\frac12,\frac{D-1}{2}\right).
\]
By symmetry of $U_1$, its density is therefore
\[
f_{U_1}(u)
=
\frac{(1-u^2)^{(D-3)/2}}
{B(1/2,(D-1)/2)},
\qquad -1<u<1.
\]
The change of variables $u=\cos\theta$ now gives
\[
f_\Theta(\theta)
=
f_{U_1}(\cos\theta)\sin\theta
=
\frac{\sin^{D-2}\theta}
{B(1/2,(D-1)/2)},
\qquad 0<\theta<\pi.
\]
Since this density is symmetric about $\pi/2$ and
\(
\cos\bigl(n(\pi-\theta)\bigr)=(-1)^n\cos(n\theta),
\)
all odd cosine moments vanish.

For the even moments, the normalizing constant cancels, so
\[
\E[\cos(2j\Theta)]
=
\frac{
\displaystyle
\int_0^\pi
\sin^{D-2}\theta\cos(2j\theta)\,d\theta
}{
\displaystyle
\int_0^\pi
\sin^{D-2}\theta\,d\theta
}.
\]
More generally, for $r>-1$ and integer $j\ge0$, set
$a=r+1$ and $b=2j$ in \cite[(5.12.6)]{watson:DLMF}.
Taking real parts and using the beta-gamma relation gives
\begin{equation}
\label{eq:cos-beta}
\int_0^\pi
\sin^r\theta\cos(2j\theta)\,d\theta
=
\frac{(-1)^j\pi\Gamma(r+1)}
{2^r
 \Gamma(r/2-j+1)
 \Gamma(r/2+j+1)}.
\end{equation}
Indeed, the phase in \cite[(5.12.6)]{watson:DLMF} becomes
$e^{i\pi j}=(-1)^j$.

Setting $r=D-2$ in \eqref{eq:cos-beta} and dividing by its
$j=0$ case yields
\[
\E[\cos(2j\Theta)]
=
(-1)^j
\frac{\Gamma(D/2)^2}
{\Gamma(D/2-j)\Gamma(D/2+j)}.
\]
Whenever the gamma factors are finite, the standard
Pochhammer-gamma relation gives
\[
(-1)^j
\frac{\Gamma(D/2)^2}
{\Gamma(D/2-j)\Gamma(D/2+j)}
=
\frac{(1-D/2)_j}{(D/2)_j},
\]
which proves \eqref{eq:supp-angular-moment}.

It remains only to note the terminating case. If $D=2q$ and
$j\ge q$, then $q-j$ is a nonpositive integer, so
$1/\Gamma(q-j)=0$.
Equivalently, the product $(1-q)_j$ contains a zero factor.
Hence the same formula gives
\[
\E[\cos(2j\Theta)]=0,
\qquad j\ge q,
\]
which is the finite termination in even dimensions.
\end{proof}

\subsection{Direct series, finite formulas, and gradients}

\begin{proof}[Proof of Theorem~\ref{thm:direct-kl}]
The transport identity in the main paper gives
\begin{equation}
\label{eq:supp-transport-kl}
K_D(\rho)
=
m\left[
-\log(1-\rho^2)
+\E\log(1+\rho^2+2\rho U_1)
\right].
\end{equation}
The distribution of $U_1$ is symmetric, so the plus sign may be
replaced by a minus sign. Since
\[
1+\rho^2-2\rho\cos\theta
=
|1-\rho e^{i\theta}|^2,
\]
the power series
\[
\log(1-z)=-\sum_{n=1}^{\infty}\frac{z^n}{n},
\qquad |z|<1,
\]
with $z=\rho e^{i\theta}$ gives
\[
\log(1+\rho^2-2\rho\cos\theta)
=
2\operatorname{Re}\log(1-\rho e^{i\theta}) =
-2\sum_{n=1}^{\infty}
\frac{\rho^n}{n}\cos(n\theta).
\]
The expansion converges uniformly in $\theta$ when $\rho$ lies in a
compact subset of $[0,1)$, so expectation and summation may be
interchanged. Lemma~\ref{lem:angular-moments} then gives
\[
\E\log(1+\rho^2-2\rho U_1)
=
-\sum_{j=1}^{\infty}
\frac{(1-D/2)_j}{(D/2)_j}\frac{\rho^{2j}}{j}.
\]
Substitution in \eqref{eq:supp-transport-kl} gives
\eqref{eq:direct-kl}.

In nonterminating dimensions, the coefficient can be written as
\begin{equation}
\label{eq:supp-gamma-coeff}
\frac{(1-D/2)_j}{(D/2)_j}\frac1j
=
\frac{\Gamma(D/2)}{\Gamma(1-D/2)}
\frac{\Gamma(j+1-D/2)}{\Gamma(j+D/2)}
\frac1j.
\end{equation}
The standard gamma-ratio asymptotic
\cite[(5.11.12)]{watson:DLMF} gives
\[
\frac{\Gamma(j+1-D/2)}{\Gamma(j+D/2)}
=
O(j^{1-D}),
\]
so \eqref{eq:supp-gamma-coeff} is $O(j^{-D})$.
This proves absolute convergence of the correction series at
$\rho=1$.
\end{proof}

\begin{proof}[Proof of Corollary~\ref{cor:positive-expansion}]
Expanding $-\log(1-\rho^2)$ in Theorem~\ref{thm:direct-kl} gives
\eqref{eq:positive-kl-series}. Write
\[
c_{D,j}
=\prod_{k=0}^{j-1}\frac{k+1-D/2}{k+D/2}.
\]
Every factor has absolute value smaller than one for $D\ge2$. Thus
$|c_{D,j}|<1$ whenever the product is nonzero, and
$1-c_{D,j}>0$. Since $0<1-c_{D,j}<2$, on every
$0\le\rho\le r$ with $0<r<1$ the first- and second-derivative
series are dominated by constant multiples of \(\sum_j r^{2j-1} \) and \(\sum_j(2j-1)r^{2j-2}\),
respectively. Both series converge, so termwise differentiation is
justified for $|\rho|<1$, giving
\eqref{eq:positive-first-derivative} and
\eqref{eq:positive-second-derivative}. In particular, $K_D'(0)=0$, $K_D'(\rho)>0$ for $0<\rho<1$, and
$K_D''(\rho)>0$ on $[0,1)$, proving strict increase and strict
convexity in $\rho$. The same positive series in $x=\rho^2$ proves
strict convexity in $x$. The first coefficient is
$2(D-1)^2/D$. Keeping this term gives
\eqref{eq:global-quadratic-lower}.
\end{proof}

\begin{proof}[Proof of Corollary~\ref{cor:even-kl} and the gradient identities]
The recurrence \eqref{eq:term-recurrence} follows from
\[
(a)_{j+1}=(a+j)(a)_j.
\]
When $D=2q$, the factor $1-q$ is a nonpositive integer, so
$(1-q)_j=0$ for $j\ge q$. This proves the finite formula in
Corollary~\ref{cor:even-kl}. The coefficient also has the form
\begin{equation}
\label{eq:combinatorial-coeff}
\frac{(1-q)_j}{(q)_j}
=(-1)^j\frac{\binom{q-1}{j}}{\binom{q+j-1}{j}},
\qquad 1\le j\le q-1.
\end{equation}
The termwise differentiation justified in the proof of
Corollary~\ref{cor:positive-expansion} gives
\eqref{eq:kl-gradient}. For the ball parameter, the chain rule gives
\[
\nabla_a K_D(\|a\|)
=
\begin{cases}
K_D'(\|a\|)a/\|a\|, & a\ne0,\\
0, & a=0.
\end{cases}
\]
The value at $a=0$ follows continuously from
$K_D(\rho)=O(\rho^2)$ as $\rho\to0$, which is immediate from
\eqref{eq:positive-kl-series}.
\end{proof}

The first finite cases are
\begin{align}
K_2(\rho)&=-\log(1-\rho^2),\\
K_4(\rho)&=3\left[-\log(1-\rho^2)+\frac{\rho^2}{2}\right],\\
K_6(\rho)&=5\left[-\log(1-\rho^2)+\frac{2\rho^2}{3}-\frac{\rho^4}{12}\right].
\end{align}

\subsection{Pairwise KL and hyperbolic distance}

\begin{proof}[Proof of Proposition~\ref{prop:pairwise-kl}]
Let $T_b$ be a M\"obius transformation that sends $b$ to $0$ and
use the same notation for its induced action on the ball parameters.
By the M\"obius closure of the spherical Cauchy family
\cite[Theorem~2.3]{spCauchy}, if $X\sim P_c$, then
\[
T_b(X)\sim P_{T_b(c)}.
\]
In particular, $T_b$ transforms $P_b$ into
$P_0=\nuD$. For the same parameter transformation,
\[
\|T_b(a)\|^2
=
\frac{\|a-b\|^2}
{1-2a^\top b+\|a\|^2\|b\|^2}
=
\delta(a,b)^2.
\]

Applying the same invertible transformation to both distributions
does not change their KL divergence. Hence
\[
\begin{aligned}
\KL(P_a\|P_b)
&=
\KL(P_{T_b(a)}\|P_0) \\
&=
K_D(\|T_b(a)\|) \\
&=
K_D(\delta(a,b)).
\end{aligned}
\]
Since $\delta(a,b)=\delta(b,a)$, the same formula also gives
\[
\KL(P_a\|P_b)=\KL(P_b\|P_a).
\]

Finally, the Poincar\'e-ball distance satisfies
\[
d_{\mathbb H}(a,b)
=
2\operatorname{atanh}\delta(a,b).
\]
Since $K_D$ is strictly increasing by
Corollary~\ref{cor:positive-expansion}, the pairwise KL is therefore
a strictly increasing function of hyperbolic distance whenever
$a\ne b$.
\end{proof}

\subsection{Certified value and gradient truncation in odd dimensions}

\begin{proof}[Proof of Corollary~\ref{cor:odd-tail}]
Let $D=2q+1$, write $x=\rho^2$, and recall that
$n=N+1\ge q$.

For the KL correction, set
\[
A_j
=
\left|
\frac{(1-D/2)_j}{(D/2)_j}
\right|\frac1j,
\]
so that $|t_j|=A_jx^j$. For $j\ge q$,
\begin{equation}
\label{eq:odd-ratio}
\frac{A_{j+1}}{A_j}
=
r_j
=
\frac{j-q+1/2}{j+q+1/2}\frac{j}{j+1}
<1.
\end{equation}
Hence
\[
\frac{|t_{j+1}|}{|t_j|}
=xr_j<x,
\]
and comparison with a geometric series gives
\begin{equation}
\label{eq:geometric-tail}
\sum_{j=n}^{\infty}|t_j|
\le
\frac{|t_n|}{1-x}.
\end{equation}

For a bound that remains finite as $x\to1$, set
\[
F_j=1+\frac{j}{2q}.
\]
Since $A_{j+1}=r_jA_j$, a direct calculation gives
\begin{equation}
\label{eq:telescoping-factor}
F_j-r_jF_{j+1}
=
1+\frac{j(2q+1)}
{(j+1)(2j+2q+1)}
>1.
\end{equation}
Therefore
\[
A_j
<
A_jF_j-A_{j+1}F_{j+1}.
\]
Summing from $j=n$ to $M$ makes the right-hand side telescope, so
\[
\sum_{j=n}^M A_j
<
A_nF_n-A_{M+1}F_{M+1}
\le A_nF_n.
\]
Letting $M\to\infty$ and using $x^j\le x^n$ for $j\ge n$ gives
\begin{equation}
\label{eq:boundary-tail}
\sum_{j=n}^{\infty}|t_j|
=
\sum_{j=n}^{\infty}A_jx^j
\le
|t_n|
\left(1+\frac{n}{2q}\right).
\end{equation}
Taking the smaller of \eqref{eq:geometric-tail} and
\eqref{eq:boundary-tail} proves \eqref{eq:odd-tail}.

For the gradient correction, set
\[
B_j
=
\left|
\frac{(1-D/2)_j}{(D/2)_j}
\right|,
\]
so that $|g_j|=B_j\rho^{2j-1}$. For $j\ge q$,
\[
\frac{B_{j+1}}{B_j}
=
s_j
=
\frac{j-q+1/2}{j+q+1/2}
<1.
\]
The same geometric argument gives
\[
\sum_{j=n}^{\infty}|g_j|
\le
\frac{|g_n|}{1-x}.
\]

For the boundary-stable bound, set
\[
G_j=1+\frac{j}{2q-1}.
\]
A direct calculation gives
\[
G_j-s_jG_{j+1}
=
\frac{2j+4q+1}{2j+2q+1}
>1.
\]
Thus, exactly as above,
\[
\sum_{j=n}^{\infty}B_j
\le
B_nG_n.
\]
Since $\rho^{2j-1}\le\rho^{2n-1}$ for $j\ge n$,
\[
\sum_{j=n}^{\infty}|g_j|
\le
|g_n|
\left(1+\frac{n}{2q-1}\right).
\]
Taking the smaller of the two bounds proves
\eqref{eq:odd-gradient-tail}.
\end{proof}

\subsection{Exact formulas for the first odd dimensions}

The adaptive recurrence is exact in every odd dimension. The first two odd cases also have compact elementary formulas.

\begin{suppprop}[Exact formulas for $D=3$ and $D=5$]
\label{prop:low-odd}
Let $z=4\rho/(1+\rho)^2$ and $L=\log((1-\rho)/(1+\rho))$. Then
\begin{align}
K_3(\rho)
&=\frac{1+\rho^2}{\rho}\log\left(\frac{1+\rho}{1-\rho}\right)-2,
\label{eq:K3}\\
K_5(\rho)
&=4\left[L+\frac{2}{z^2}-\frac{2}{z}-\frac56
+\frac{2-3z}{z^3}\log(1-z)\right].
\label{eq:K5}
\end{align}
Both extend continuously to $K_D(0)=0$.
\end{suppprop}

\begin{proof}
Use the compact integral representation \eqref{eq:integral-kl}.

For $D=3$,
\[
J_3(z)
=
\int_0^1
\frac{t}{1-t}
\left(
1-\frac{1-z}{1-zt}
\right)dt.
\]
Since
\[
1-\frac{1-z}{1-zt}
=
\frac{z(1-t)}{1-zt},
\]
the factor $1-t$ cancels and
\[
J_3(z)
=
z\int_0^1\frac{t}{1-zt}\,dt.
\]
Using
\[
\frac{zt}{1-zt}
=
-1+\frac{1}{1-zt},
\]
we obtain
\[
J_3(z)
=
\int_0^1
\left(
-1+\frac{1}{1-zt}
\right)dt
=
-1-\frac{\log(1-z)}{z}.
\]
Since $\log(1-z)=2L$, substitution in
\eqref{eq:integral-kl} gives \eqref{eq:K3}.

For $D=5$,
\[
J_5(z)
=
\int_0^1
\frac{t^3}{1-t}
\left[
1-
\left(
\frac{1-z}{1-zt}
\right)^2
\right]dt.
\]
Here
\[
(1-zt)^2-(1-z)^2
=
z(1-t)\bigl(2-z(1+t)\bigr),
\]
so again the factor $1-t$ cancels and
\[
J_5(z)
=
z\int_0^1
\frac{t^3\bigl(2-z(1+t)\bigr)}
{(1-zt)^2}\,dt.
\]
The integrand has the partial-fraction decomposition
\[
z\frac{t^3\bigl(2-z(1+t)\bigr)}
{(1-zt)^2}
=
-t^2-t-\frac{2z-1}{z^2}
+\frac{3z-2}{z^2(1-zt)}
-\frac{z-1}{z^2(1-zt)^2}.
\]
Integrating term by term and using
\[
\int_0^1\frac{dt}{1-zt}
=
-\frac{\log(1-z)}{z},
\qquad
\int_0^1\frac{dt}{(1-zt)^2}
=
\frac{1}{1-z},
\]
gives
\[
J_5(z)
=
\frac{2}{z^2}-\frac{2}{z}-\frac56
+\frac{2-3z}{z^3}\log(1-z).
\]
Substitution in \eqref{eq:integral-kl} gives \eqref{eq:K5}.
Taylor expansion of $\log(1-z)$ at $z=0$ shows that the apparent
singularities cancel and proves the continuous limits at $\rho=0$.
\end{proof}

\subsection{Hypergeometric and compact integral representations}

The direct series is the primary exact evaluator. We derive the second
representation because it is useful for monotonicity, asymptotic
analysis, and analytic bounds.

\begin{proof}[Proof of Proposition~\ref{prop:integral}]
For $\gamma>0$, define
\begin{equation}
\label{eq:Zgamma}
Z(\gamma)
=
\int_{\SD}
\frac{(1-\rho^2)^m}
{(1+\rho^2-2\rho\mu^\top x)^\gamma}
\,d\nuD(x).
\end{equation}
At $\gamma=m$ the integrand is the spherical Cauchy density, so
$Z(m)=1$. Differentiating under the integral sign therefore gives
\begin{equation}
\label{eq:Z-derivative}
\E_q\log(1+\rho^2-2\rho\mu^\top X)
=
-\left.
\frac{d}{d\gamma}\log Z(\gamma)
\right|_{\gamma=m}.
\end{equation}

To evaluate $Z(\gamma)$, let $U\sim\nuD$ and set
\[
Y=\frac{1+\mu^\top U}{2}.
\]
The uniform-sphere marginal gives
\[
Y\sim\operatorname{Beta}\left(\frac m2,\frac m2\right).
\]
With
\[
z=\frac{4\rho}{(1+\rho)^2},
\]
we have
\[
1+\rho^2-2\rho\mu^\top U
=
(1+\rho)^2(1-zY).
\]
Hence Euler's integral representation of the Gauss hypergeometric
function \cite[(15.6.1)]{watson:DLMF}, together with the normalization
in \cite[(15.1.2)]{watson:DLMF}, gives the exact identity
\begin{equation}
\label{eq:Z-hypergeo}
Z(\gamma)
=
(1-\rho^2)^m(1+\rho)^{-2\gamma}
{}_2F_1\left(\gamma,\frac m2;m;z\right).
\end{equation}

It remains to differentiate the hypergeometric factor. For $z<1$,
the Gauss series \cite[(15.2.1)]{watson:DLMF} gives
\[
{}_2F_1\left(\gamma,\frac m2;m;z\right)
=
\sum_{k=0}^{\infty}
\frac{(\gamma)_k(m/2)_k}{(m)_k\,k!}z^k.
\]
The series is locally uniformly convergent in $\gamma$ near $m$, so it
may be differentiated termwise. Using
\[
\frac{d}{d\gamma}(\gamma)_k
=
(\gamma)_k
\bigl[\psi(\gamma+k)-\psi(\gamma)\bigr],
\]
which follows from
\cite[(5.2.2), (5.2.5)]{watson:DLMF}, and then setting
$\gamma=m$, we obtain
\[
\left.
\frac{d}{d\gamma}
{}_2F_1\left(\gamma,\frac m2;m;z\right)
\right|_{\gamma=m}
=
\sum_{k=0}^{\infty}
\frac{(m/2)_k}{k!}z^k
\bigl[\psi(m+k)-\psi(m)\bigr].
\]
Moreover,
\[
{}_2F_1\left(m,\frac m2;m;z\right)
=
(1-z)^{-m/2}
\]
by \cite[(15.4.6)]{watson:DLMF}. Dividing the derivative by this
value gives
\begin{equation}
\label{eq:hypergeo-log-derivative}
\left.
\frac{d}{d\gamma}
\log{}_2F_1\left(\gamma,\frac m2;m;z\right)
\right|_{\gamma=m}
=
(1-z)^{m/2}
\sum_{k=0}^{\infty}
\frac{(m/2)_k}{k!}z^k
\bigl[\psi(m+k)-\psi(m)\bigr].
\end{equation}

To convert this series to an integral, subtract
\cite[(5.9.16)]{watson:DLMF} evaluated at $m+k$ and at $m$:
\[
\psi(m+k)-\psi(m)
=
\int_0^1
\frac{t^{m-1}(1-t^k)}{1-t}\,dt.
\]
Substitution in \eqref{eq:hypergeo-log-derivative} gives
\[
(1-z)^{m/2}
\int_0^1
\frac{t^{m-1}}{1-t}
\sum_{k=0}^{\infty}
\frac{(m/2)_k}{k!}z^k(1-t^k)\,dt.
\]
All summands are nonnegative for $0\le z<1$ and $0\le t\le1$, so
the sum and integral may be interchanged. The binomial identity
\cite[(15.4.6)]{watson:DLMF} gives
\[
\sum_{k=0}^{\infty}
\frac{(m/2)_k}{k!}z^k
=
(1-z)^{-m/2},
\]
and likewise
\[
\sum_{k=0}^{\infty}
\frac{(m/2)_k}{k!}(zt)^k
=
(1-zt)^{-m/2}.
\]
Therefore
\begin{align}
\left.
\frac{d}{d\gamma}
\log{}_2F_1\left(\gamma,\frac m2;m;z\right)
\right|_{\gamma=m}
&=
\int_0^1
\frac{t^{m-1}}{1-t}
\left[
1-
\left(
\frac{1-z}{1-zt}
\right)^{m/2}
\right]dt \\
&=J_D(z).
\label{eq:supp-J}
\end{align}

Finally, differentiating \eqref{eq:Z-hypergeo} with respect to
$\gamma$ gives
\[
\left.
\frac{d}{d\gamma}\log Z(\gamma)
\right|_{\gamma=m}
=
-2\log(1+\rho)+J_D(z).
\]
Together with \eqref{eq:Z-derivative},
\[
\E_q\log(1+\rho^2-2\rho\mu^\top X)
=
2\log(1+\rho)-J_D(z).
\]
Hence
\[
\begin{aligned}
K_D(\rho)
&=
m\left[
\log(1-\rho^2)
-
\E_q\log(1+\rho^2-2\rho\mu^\top X)
\right] \\
&=
m\left[
\log\left(\frac{1-\rho}{1+\rho}\right)
+
J_D(z)
\right],
\end{aligned}
\]
which proves \eqref{eq:integral-kl}.
\end{proof}

\subsection{Asymptotic formulas}

\begin{proof}[Proof of Proposition~\ref{prop:asymptotics}] 
The first direct-series coefficient is
\[
\frac{1-D/2}{D/2}\rho^2
=-\frac{D-2}{D}\rho^2.
\]
Together with $-\log(1-\rho^2)=\rho^2+O(\rho^4)$, this gives
\[
K_D(\rho)
=(D-1)\left(1+\frac{D-2}{D}\right)\rho^2+O(\rho^4)
=\frac{2(D-1)^2}{D}\rho^2+O(\rho^4).
\]

For $\rho\to1^-$, use the compact representation
\eqref{eq:integral-kl}. Put
\[
\delta=\frac{m}{2},
\qquad
z=\frac{4\rho}{(1+\rho)^2}.
\]
We first show that
\begin{equation}
\label{eq:J-near-one}
J_D(z)
=
-\log(1-z)
+\psi(\delta)-\psi(m)
+o(1),
\qquad z\to1^-.
\end{equation}

Using
\[
\frac{1-t^{m-1}}{1-t}
=
\sum_{\lambda=0}^{m-2}t^\lambda,
\]
with the sum empty when $m=1$, write
\[
J_D(z)=J_1(z)-J_2(z),
\]
where
\[
J_1(z)
=
\int_0^1
\frac{1}{1-t}
\left[
1-\left(\frac{1-z}{1-zt}\right)^\delta
\right]dt
\]
and
\[
J_2(z)
=
\sum_{\lambda=0}^{m-2}
\int_0^1
t^\lambda
\left[
1-\left(\frac{1-z}{1-zt}\right)^\delta
\right]dt.
\]

For $J_1$, set
\[
a=\frac{1-z}{1-zt}.
\]
Since
\[
\frac{1-a}{1-t}
=
\frac{z}{1-zt},
\]
we obtain
\[
J_1(z)
=
\int_0^1\frac{z}{1-zt}\,dt
+
z\int_0^1
\frac{1}{1-zt}
\frac{a-a^\delta}{1-a}\,dt.
\]
The first integral is $-\log(1-z)$. In the second, the substitution
$a=e^{-u}$ gives
\[
J_1(z)
=
-\log(1-z)
+
\int_0^{-\log(1-z)}
\frac{e^{-u}-e^{-\delta u}}
{1-e^{-u}}\,du.
\]
The last integrand is bounded near $u=0$ and decays exponentially as
$u\to\infty$, so the integral converges absolutely. Subtracting
\cite[(5.9.12)]{watson:DLMF} at $\delta$ and at $1$ therefore gives
\[
J_1(z)
=
-\log(1-z)
+\psi(\delta)-\psi(1)
+o(1).
\]

For $J_2$, the bracket lies in $[0,1]$ and converges to $1$ for every
$t<1$ as $z\to1^-$. Dominated convergence thus gives
\[
J_2(z)
\longrightarrow
\sum_{\lambda=0}^{m-2}\int_0^1t^\lambda\,dt
=
\sum_{\lambda=1}^{m-1}\frac1\lambda
=
\psi(m)-\psi(1),
\]
where the last equality follows from the digamma recurrence
\cite[(5.5.2)]{watson:DLMF}. This proves
\eqref{eq:J-near-one}.

Finally,
\[
1-z
=
\frac{(1-\rho)^2}{(1+\rho)^2},
\]
so
\[
-\log(1-z)
=
2\log\left(\frac{1+\rho}{1-\rho}\right).
\]
Substitution in \eqref{eq:integral-kl} yields
\[
K_D(\rho)
=
m\left[
\log\left(\frac{1+\rho}{1-\rho}\right)
+\psi\left(\frac m2\right)-\psi(m)
\right]
+o(1).
\]
Since
\[
\log\left(\frac{1+\rho}{1-\rho}\right)
=
-\log(1-\rho^2)+2\log(1+\rho)
=
-\log(1-\rho^2)+2\log2+o(1),
\]
we obtain \eqref{eq:high-rho}.

For the high-dimensional limit, set
\[
c_{D,j}=\frac{(1-D/2)_j}{(D/2)_j}.
\]
For fixed $j$, $c_{D,j}\to(-1)^j$ as $D\to\infty$. The product representation also gives $|c_{D,j}|\le1$. For fixed $\rho<1$, dominated convergence in \eqref{eq:direct-kl} yields
\begin{align*}
\lim_{D\to\infty}\frac{K_D(\rho)}{D-1}
&=-\log(1-\rho^2)-\sum_{j=1}^{\infty}(-1)^j\frac{\rho^{2j}}{j}\\
&=\log\left(\frac{1+\rho^2}{1-\rho^2}\right),
\end{align*}
which proves \eqref{eq:large-D}.
\end{proof}

\subsection{Monotonicity in ambient dimension}

\begin{proof}[Proof of Proposition~\ref{prop:dimension-monotonicity}] 
Fix $z\in(0,1)$ and set $\delta=(D-1)/2$. Let
\[
A(t)=\frac{1-z}{1-zt}.
\]
From \eqref{eq:J-def}, write
\[
J_{D+1}(z)-J_D(z)=I_1+I_2,
\]
where
\[
I_1
=\int_0^1\frac{t^{D-1}-t^{D-2}}{1-t}\,dt
=-\frac{1}{D-1}
\]
and
\[
I_2
=\int_0^1\frac{t^{D-2}}{1-t}
\left[A(t)^\delta-tA(t)^{\delta+1/2}\right]dt.
\]
For $0<t<1$,
\begin{align*}
&1-zt-t\sqrt{1-z}\sqrt{1-zt}
-
\left(1-\frac{zt}{2}\right)(1-t)\\
&\hspace{25mm}
=\frac{t}{2}\left(\sqrt{1-zt}-\sqrt{1-z}\right)^2>0.
\end{align*}
It follows that
\[
\frac{1}{1-t}
\left[
\frac{1}{(1-zt)^\delta}
-
\frac{t\sqrt{1-z}}{(1-zt)^{\delta+1/2}}
\right]
>
\frac{1-zt/2}{(1-zt)^{\delta+1}}.
\]
Multiplying by $t^{D-2}(1-z)^\delta$ and integrating gives
\begin{align*}
I_2
&>
(1-z)^\delta
\int_0^1
\frac{t^{D-2}(1-zt/2)}{(1-zt)^{\delta+1}}dt\\
&=
\frac{(1-z)^\delta}{D-1}
\left[
\frac{t^{D-1}}{(1-zt)^\delta}
\right]_{0}^{1}
=
\frac{1}{D-1}.
\end{align*}
Hence $J_{D+1}(z)>J_D(z)$. Since the logarithmic term in \eqref{eq:integral-kl} does not depend on $D$,
\[
\frac{K_{D+1}(\rho)}{D}
>
\frac{K_D(\rho)}{D-1}.
\]
The normalized KL is positive for $0<\rho<1$. Therefore
\[
K_{D+1}(\rho)
>
\frac{D}{D-1}K_D(\rho)
>
K_D(\rho).
\]
\end{proof}

\subsection{Even-neighbor approximation in odd dimensions}

\begin{proof}[Proof of Corollary~\ref{cor:neighbor-average}]
Let $D\ge3$ be odd. Proposition~\ref{prop:dimension-monotonicity} gives
\[
K_{D-1}(\rho)\le K_D(\rho)\le K_{D+1}(\rho),
\]
with strict inequalities for $0<\rho<1$. Any point in an interval differs from its center by at most half the interval width. Hence
\[
\left|
\frac{K_{D-1}(\rho)+K_{D+1}(\rho)}{2}
-K_D(\rho)
\right|
\le
\frac{K_{D+1}(\rho)-K_{D-1}(\rho)}{2},
\]
which proves \eqref{eq:neighbor-certificate}.

Both neighboring dimensions are even. Substituting \eqref{eq:finite-even-kl} gives
\begin{align*}
K_D^{\mathrm{nbr}}(\rho)
={}&\frac{D-2}{2}\left[
-\log(1-\rho^2)
-\sum_{j\ge1}c_{D-1,j}\frac{\rho^{2j}}{j}
\right]\\
&+\frac{D}{2}\left[
-\log(1-\rho^2)
-\sum_{j\ge1}c_{D+1,j}\frac{\rho^{2j}}{j}
\right].
\end{align*}
The coefficient of the logarithm is
\[
\frac{D-2+D}{2}=D-1.
\]
After factoring out $D-1$, the coefficient of $\rho^{2j}/j$ is
\[
\overline c_{D,j}
=\frac{(D-2)c_{D-1,j}+Dc_{D+1,j}}{2(D-1)}.
\]
The $D-1$ series terminates after $(D-3)/2$ terms, and the $D+1$ series terminates after $(D-1)/2$ terms. Setting the missing coefficient in the shorter series to zero proves \eqref{eq:neighbor-polynomial} and \eqref{eq:neighbor-coefficients}.
\end{proof}

The centered average also explains the accuracy of the construction. Any term in $K_D(\rho)$ that is affine in $D$ is reproduced exactly by \eqref{eq:neighbor-average}. The leading large-dimensional term in \eqref{eq:large-D} has this form. The approximation therefore acts on the smaller nonlinear remainder in the dimension variable.

\subsection{Analytic KL envelopes}

\begin{proof}[Proof of Proposition~\ref{prop:bracket}]
Let
\[
\delta=\frac{m}{2},
\qquad
H_D(z)=J_D(z)+\log(1-z).
\]
On every compact interval $0\le z\le r<1$, the derivative of the
integrand defining $J_D$ is bounded by a constant multiple of
$t^{m-1}$. Differentiation under the integral sign is therefore
justified and gives
\begin{equation}
\label{eq:Jprime}
J_D'(z)
=
\delta(1-z)^{\delta-1}
\int_0^1
t^{2\delta-1}(1-zt)^{-\delta-1}\,dt.
\end{equation}
After the substitution $r=zt$,
\[
(1-z)J_D'(z)
=
\delta
\left(\frac{1-z}{z^2}\right)^\delta
\int_0^z
r^{2\delta-1}(1-r)^{-\delta-1}\,dr.
\]
Set
\[
h(r)=\left(\frac{r^2}{1-r}\right)^\delta.
\]
Since
\[
h'(r)
=
\delta r^{2\delta-1}(1-r)^{-\delta-1}(2-r),
\]
we obtain
\[
(1-z)J_D'(z)
=
\left(\frac{1-z}{z^2}\right)^\delta
\int_0^z\frac{h'(r)}{2-r}\,dr.
\]
For $0\le r\le z$,
\[
\frac{1}{2-r}\le\frac{1}{2-z},
\]
and therefore
\[
\begin{aligned}
(1-z)J_D'(z)
&\le
\left(\frac{1-z}{z^2}\right)^\delta
\frac{1}{2-z}
\int_0^z h'(r)\,dr \\
&=
\left(\frac{1-z}{z^2}\right)^\delta
\frac{h(z)}{2-z}
=
\frac{1}{2-z}.
\end{aligned}
\]
Consequently,
\begin{equation}
\label{eq:Hprime-bound}
H_D'(z)
=
J_D'(z)-\frac{1}{1-z}
\le
-\frac{1}{2-z}.
\end{equation}

Since $H_D(0)=0$, integration from $0$ to $z$ gives
\begin{equation}
\label{eq:Hupper}
H_D(z)
\le
-\int_0^z\frac{du}{2-u}
=
\log\left(1-\frac z2\right).
\end{equation}
On the other hand, the high-concentration expansion
\eqref{eq:J-near-one} gives
\[
H_D(1^-)
=
\psi(\delta)-\psi(m).
\]
Integrating \eqref{eq:Hprime-bound} backward from $1$ yields
\begin{equation}
\label{eq:Hlower}
H_D(z)
\ge
\psi(\delta)-\psi(m)+\log(2-z).
\end{equation}

Since
\[
1-z
=
\left(\frac{1-\rho}{1+\rho}\right)^2,
\]
the compact representation can be written as
\[
\frac{K_D(\rho)}{m}
=
\log\left(\frac{1+\rho}{1-\rho}\right)
+
H_D(z).
\]
Moreover,
\[
\frac{1+\rho}{1-\rho}
\left(1-\frac z2\right)
=
\frac{1+\rho^2}{1-\rho^2},
\]
and
\[
\frac{1+\rho}{1-\rho}(2-z)
=
\frac{2(1+\rho^2)}{1-\rho^2}.
\]
Substituting \eqref{eq:Hupper} and \eqref{eq:Hlower} therefore gives
\eqref{eq:upper-bound} and \eqref{eq:lower-bound}, respectively.
The bounds extend to $\rho=0$ by continuity.

Their difference is
\[
m\left[
\psi(m)-\psi\left(\frac m2\right)-\log2
\right]
=
mw_D.
\]
Finally, the standard digamma expansion
\cite[(5.11.2)]{watson:DLMF} gives
\[
w_D
=
\frac{1}{2m}
+\frac{1}{4m^2}
+O(m^{-3}),
\]
and hence
\[
mw_D
=
\frac12+O(m^{-1})
=
\frac12+O(D^{-1}).
\]
\end{proof}

\subsection{Large-dimensional origin of the Laplace weight}

With $m=D-1$,
\begin{equation}
\label{eq:c-m-product}
c_{D,j}
=\frac{((1-m)/2)_j}{((m+1)/2)_j}
=(-1)^j\prod_{k=0}^{j-1}\frac{m-1-2k}{m+1+2k}.
\end{equation}
For fixed $j$,
\begin{equation}
\label{eq:c-expansion}
c_{D,j}
=(-1)^j\left(1-\frac{2j^2}{m}+O(m^{-2})\right).
\end{equation}
For fixed $\rho<1$, geometric decay permits termwise summation. Set $x=\rho^2$. Then
\begin{align}
\frac{K_D(\rho)}{m}
&=\log\left(\frac{1+x}{1-x}\right)
+\frac{2}{m}\sum_{j=1}^{\infty}(-1)^j jx^j+O(m^{-2})\\
&=\log\left(\frac{1+x}{1-x}\right)
-\frac{2x}{m(1+x)^2}+O(m^{-2}).
\label{eq:K-large-second}
\end{align}
On the other hand, the weight in \eqref{eq:laplace-weight} satisfies
\[
w_D\omega(\rho)
=w_D\left(\frac{2\rho}{1+\rho^2}\right)^2
=\left(\frac{1}{2m}+O(m^{-2})\right)\frac{4x}{(1+x)^2}
=\frac{2x}{m(1+x)^2}+O(m^{-2}).
\]
Thus the Laplace-weighted surrogate matches \eqref{eq:K-large-second} through first order. Multiplying the remaining $O(m^{-2})$ term by $m$ proves \eqref{eq:surrogate-asymptotic}.

\section{Numerical evaluation}
\label{app:numerics}

\subsection{Matched-curvature illustration}

\begin{figure}[htbp]
    \centering
    \includegraphics[width=0.72\textwidth]{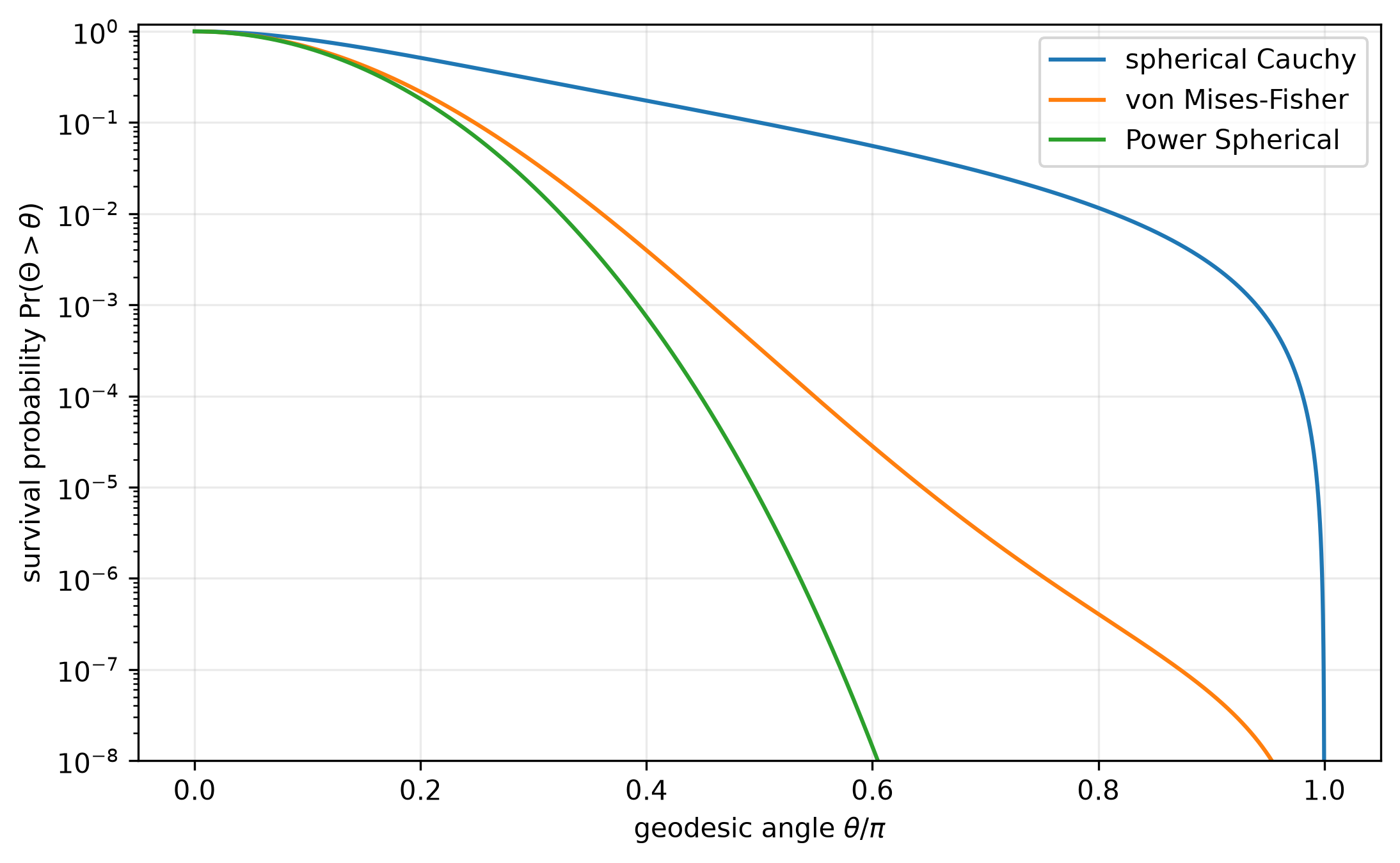}
    \caption{Survival probabilities of the geodesic angle for $D=3$ and matched curvature $\kappa=8$. Spherical Cauchy uses $\rho=0.5$, and the Power Spherical exponent is $16$.}
    \label{fig:supp-matched-survival}
\end{figure}

Figure~\ref{fig:supp-matched-survival} illustrates the angular-tail ordering in Proposition~\ref{prop:profile-order}. The three distributions have the same quadratic curvature at the mode, yet their survival probabilities separate over the full angular range.

\subsection{KL formulas for the matched-curvature comparison}

For completeness, we record the KL formulas used in the right panel of the matched-curvature figure in the main paper. Let
\[
A_{D-1}=\frac{2\pi^{D/2}}{\Gamma(D/2)}
\]
be the surface area of $\mathbb S^{D-1}$. For vMF,
\begin{equation}
\label{eq:supp-vmf-kl}
\KL\!\left(\vMF_D(\mu,\kappa)\middle\|\nuD\right)
=\log\!\left[A_{D-1}C_D(\kappa)\right]
+\kappa\frac{I_{D/2}(\kappa)}{I_{D/2-1}(\kappa)},
\end{equation}
where
\[
C_D(\kappa)
=\frac{\kappa^{D/2-1}}{(2\pi)^{D/2}I_{D/2-1}(\kappa)}.
\]

For Power Spherical, let $a=(D-1)/2$ and let $\lambda$ be the exponent in $(1+\mu^\top x)^\lambda$. Under this law,
\[
T=\frac{1+\mu^\top X}{2}
\sim\operatorname{Beta}(a+\lambda,a).
\]
Its KL to the uniform prior is
\begin{equation}
\label{eq:supp-ps-kl}
\KL\!\left(\PS_D(\mu,\lambda)\middle\|\nuD\right)
=\log\frac{\mathrm B(a,a)}{\mathrm B(a+\lambda,a)}
+\lambda\left[\psi(a+\lambda)-\psi(2a+\lambda)\right].
\end{equation}
The matched-curvature plot uses $\lambda=2\kappa$ and the stable spherical Cauchy inversion
\[
\rho
=\frac{\kappa}{m+\kappa+\sqrt{m^2+2m\kappa}},
\qquad m=D-1.
\]

\subsection{Evaluator error grid}

\begin{figure}[htbp]
\centering
\includegraphics[width=\textwidth]{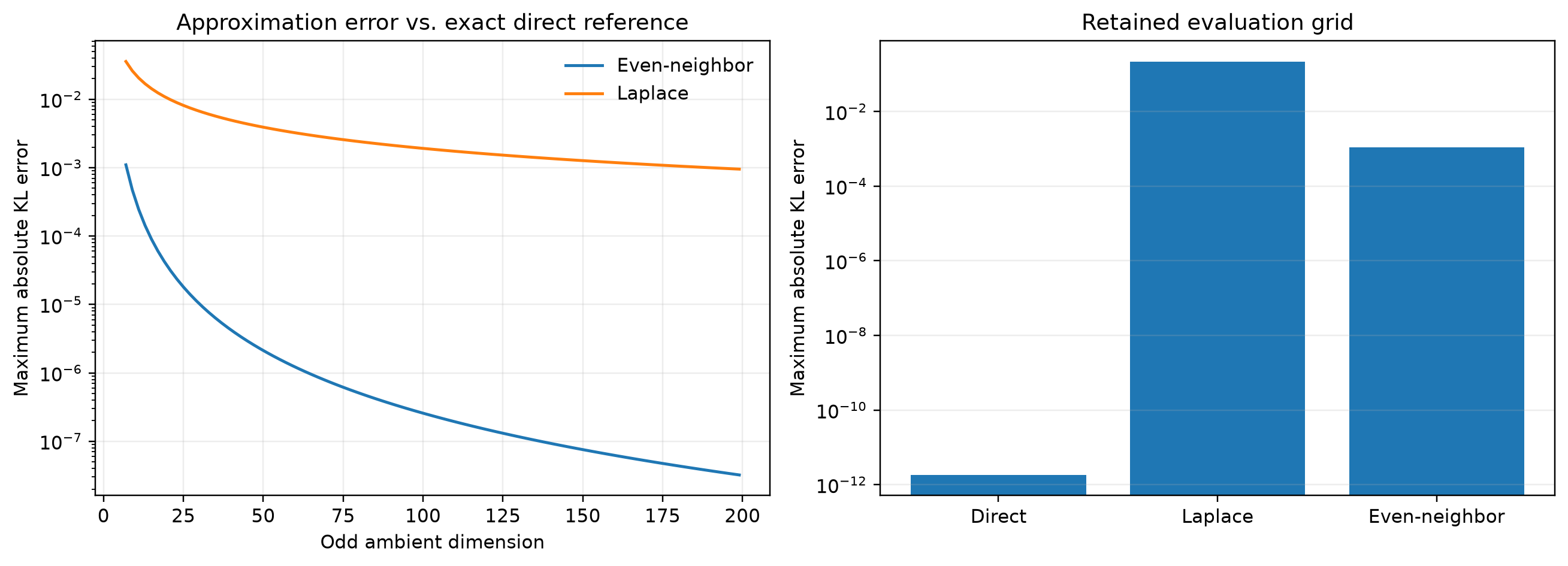}
\caption{Accuracy of the KL evaluation routes. The left panel shows the maximum error over the test grid at each concentration. The right panel compares the worst even-neighbor and Laplace-weighted errors in odd dimensions.}
\label{fig:supp-kl-diagnostics}
\end{figure}

Figure~\ref{fig:supp-kl-diagnostics} uses the direct recurrence as the reference. The plotted value at each $\rho$ is the maximum over
\[
D\in\{3,5,7,8,9,16,17,32,33,64,65,128,129,256,257,512,513,1024,1025,2048\}.
\]
The finite rule \eqref{eq:finite-rule} is exact for every even $D$ and for $D=3,5$. Its nonzero error comes only from the even-neighbor approximation in odd dimensions $D\ge7$. The Laplace-weighted surrogate is evaluated in every dimension. The right panel summarizes the retained direct, even-neighbor, and Laplace routes over the complete grid.

For the right panel, the script numerically maximizes the approximation errors over $\rho\in[0,1)$ for every odd dimension from $7$ to $199$. A dense grid locates candidate interior maxima, bounded scalar optimization refines them, and the exact high-concentration constant supplies the boundary candidate. Table~\ref{tab:neighbor-errors} records representative dimensions.

\begin{table}[t]
\centering
\small
\caption{Numerically maximized absolute KL error of the even-neighbor and Laplace-weighted approximations in representative odd dimensions.}
\label{tab:neighbor-errors}
\begin{tabular}{rrr}
\toprule
$D$ & Even-neighbor error & Laplace-weighted error \\
\midrule
7   & $1.09950\times10^{-3}$ & $3.56416\times10^{-2}$ \\
9   & $4.73937\times10^{-4}$ & $2.60372\times10^{-2}$ \\
11  & $2.45199\times10^{-4}$ & $2.04619\times10^{-2}$ \\
21  & $3.10954\times10^{-5}$ & $9.82529\times10^{-3}$ \\
51  & $1.99840\times10^{-6}$ & $3.82391\times10^{-3}$ \\
101 & $2.49950\times10^{-7}$ & $1.89362\times10^{-3}$ \\
\bottomrule
\end{tabular}
\end{table}

The largest observed even-neighbor error is \(0.00109950005268\)
at $D=7$ as $\rho\to1^-$. The largest Laplace-weighted error is \(0.0356415731933\) at $D=7$ and $\rho\approx0.455907$. The worst observed error is therefore smaller by a factor of $\approx 32.4$ for the even-neighbor rule. The ratio increases over the subsequent odd dimensions. 

\subsection{Certified term counts in odd dimensions}

\begin{figure}[htbp]
    \centering
    \includegraphics[width=0.88\textwidth]{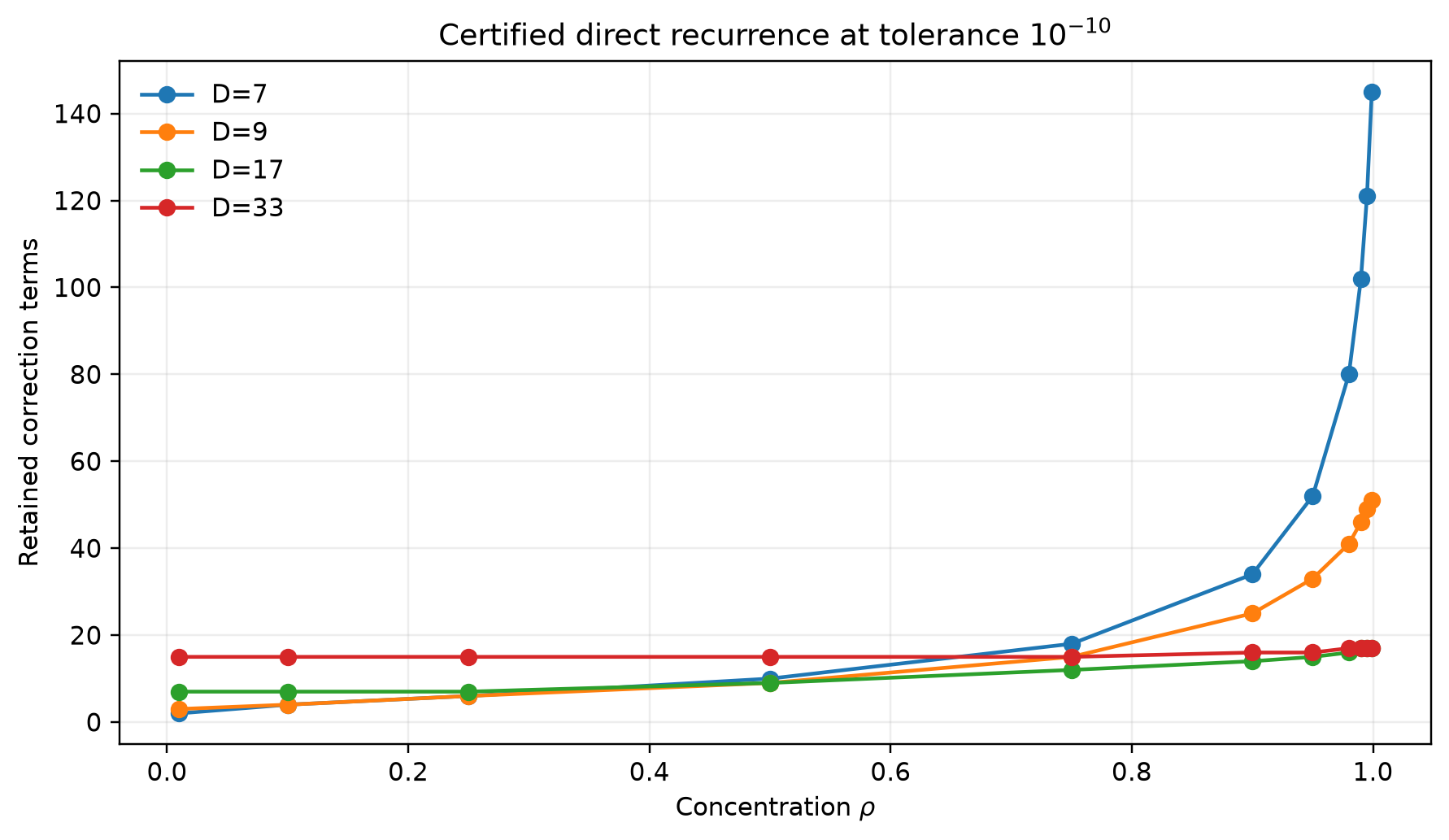}
    \caption{Terms retained by the certified odd-dimensional recurrence when the absolute KL and gradient tolerances are both $10^{-10}$. The stopping rule requires both certificates in Corollary~\ref{cor:odd-tail} to pass.}
    \label{fig:supp-odd-terms}
\end{figure}

Figure~\ref{fig:supp-odd-terms} shows the number of terms required when both the absolute KL and absolute concentration-gradient tolerances are $10^{-10}$ under the stopping rule \eqref{eq:odd-tail}. The boundary-stable parts of the two bounds prevent a blow-up as $\rho\to1^-$. At $\rho=0.999$, the rule retains $145$ terms for $D=7$, $51$ for $D=9$, $17$ for $D=17$, and $17$ for $D=33$. The low cases $D=3,5$ instead use the elementary formulas in Proposition~\ref{prop:low-odd}.

\section{Latent-layer benchmark details}
\label{app:latent-benchmark}

The benchmark times mean-direction generation, latent sampling, KL evaluation, scalar loss construction, and backward propagation. The reported comparison uses the full direct recurrence with analytic custom backward, the finite even-neighbor approximation, and the Laplace approximation. It also includes original and robust vMF implementations and the official Power Spherical path.

Runtime uses float32, $10$ warm-up iterations, $50$ measured iterations in each of five repeats, CPU batch size $128$, and GPU batch size $1024$. CUDA is synchronized around each timed component. Concentrations are matched at $\kappa=10$. Measurements were made on an Intel Core i7-10875H and an NVIDIA GeForce RTX 2070 Max-Q with 8 GiB, using Python 3.12.13, PyTorch 2.11.0 with CUDA 12.8, and Triton 3.6.0.

At $D=128$, direct spherical Cauchy requires $0.925$ ms on CPU and $1.481$ ms on CUDA. Power Spherical requires $1.347$ and $2.335$ ms, while robust vMF requires $3.846$ and $7.945$ ms.

\begin{figure}[htbp]
    \centering
    \includegraphics[width=\textwidth]{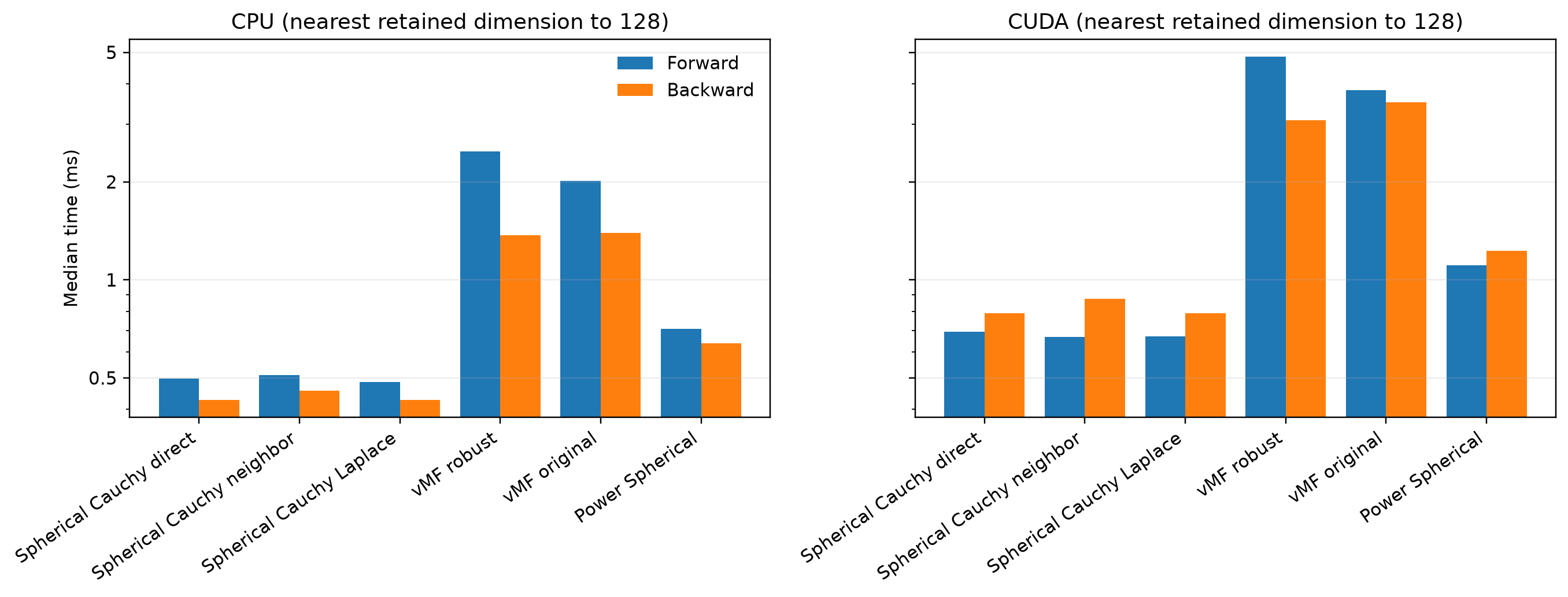}
    \caption{Forward and backward runtime decomposition for the latent-layer benchmark.}
    \label{fig:supp-runtime-split}
\end{figure}

\begin{figure}[htbp]
    \centering
    \includegraphics[width=\textwidth]{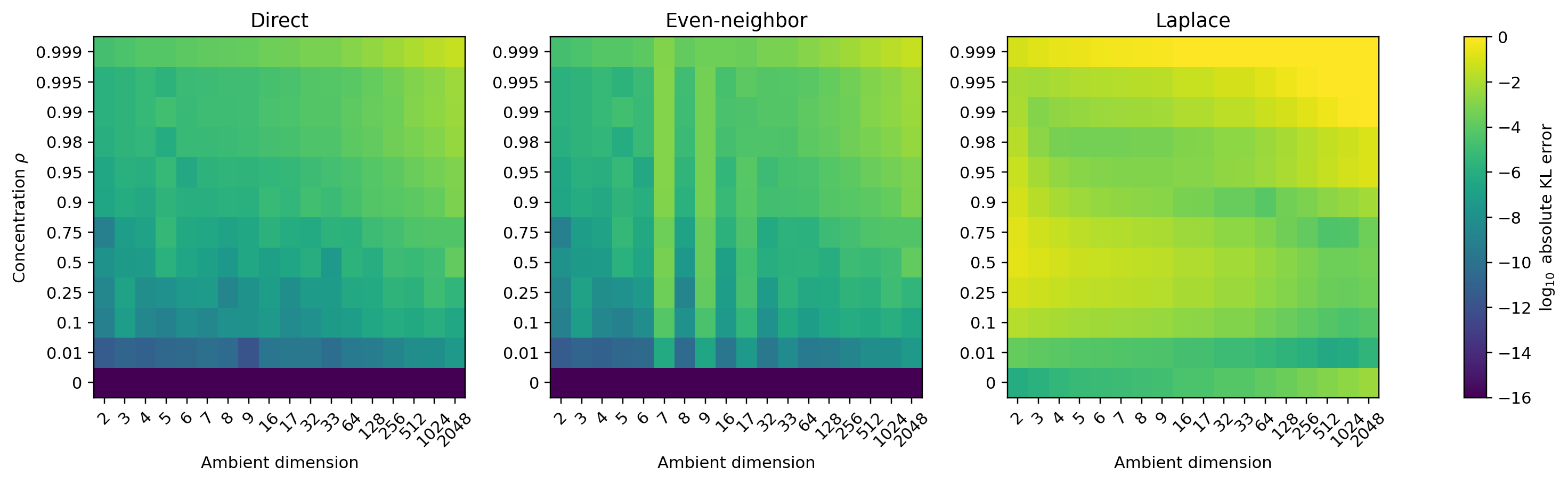}\\[2mm]
    \includegraphics[width=\textwidth]{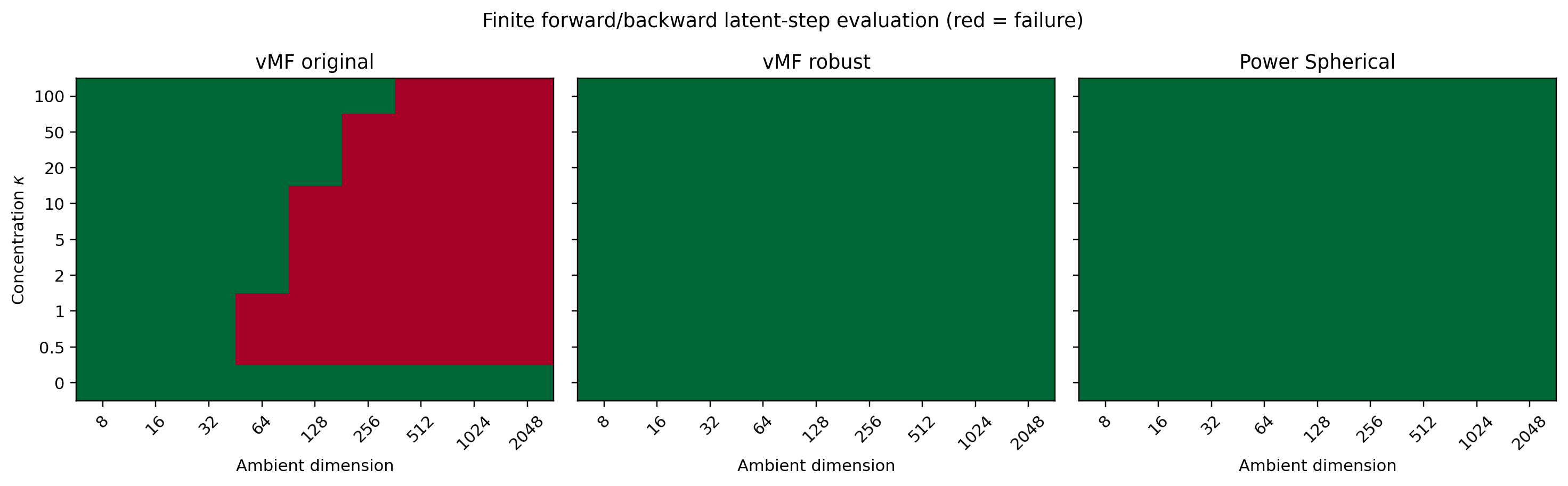}
    \caption{Robustness sweeps. The upper row shows $\log_{10}$ absolute KL error relative to the certified direct reference, darker colors indicate smaller error. Only the lower row uses green and red: green marks finite forward and backward losses and gradients, while red marks a failed run.}
    \label{fig:supp-robustness}
\end{figure}

\section{MNIST experiment details}
\label{app:mnist}

Gaussian, vMF, spherical Cauchy, and Power Spherical posteriors share a CNN with hidden dimensions $[32,64,128]$, ReLU activations, AdamW, a ReduceLROnPlateau scheduler, batch size $128$, and KL weight $1$. Training uses the standard MNIST training split and the official test split. For $p=2,3$, the learning rate is $3\times10^{-4}$ with $200$ warm-up optimizer steps. For $p\in\{5,10,20\}$, the learning rate is $10^{-4}$ without warm-up. Every run uses $40$ epochs and seeds $0$ through $4$ with deterministic data ordering. The selected checkpoint has the lowest held-out reconstruction loss.

\begin{table}[t]
\centering
\caption{Complete MNIST results. Values are means and standard deviations over five seeds. Lower is better. A dagger marks a paired comparison with the next-lowest method after Benjamini-Hochberg correction across dimensions.}
\label{tab:supp-mnist-full}
\small
\setlength{\tabcolsep}{6.5pt}
\renewcommand{\arraystretch}{1.08}
\begin{tabular}{lcccc}
\toprule
Model & $p$ & Reconstruction & Total & KL \\
\midrule
Gaussian & 2  & \ms{133.07}{0.59} & \ms{140.00}{0.60} & \ms{6.93}{0.07} \\
spCauchy & 2  & \bmsd{130.29}{0.40} & \ms{137.95}{0.41} & \ms{7.67}{0.04} \\
vMF      & 2  & \ms{132.03}{1.32} & \ms{138.84}{1.25} & \ms{6.81}{0.07} \\
Power    & 2  & \ms{132.53}{0.82} & \ms{139.29}{0.71} & \ms{6.76}{0.10} \\
\midrule
Gaussian & 3  & \ms{118.65}{1.07} & \ms{127.80}{1.11} & \ms{9.15}{0.12} \\
spCauchy & 3  & \bmsd{116.44}{0.63} & \ms{126.29}{0.62} & \ms{9.85}{0.05} \\
vMF      & 3  & \ms{118.25}{0.37} & \ms{127.29}{0.36} & \ms{9.04}{0.03} \\
Power    & 3  & \ms{119.04}{0.48} & \ms{128.00}{0.45} & \ms{8.96}{0.05} \\
\midrule
Gaussian & 5  & \ms{102.40}{1.25} & \ms{114.92}{1.16} & \ms{12.53}{0.13} \\
spCauchy & 5  & \bmsd{100.22}{0.21} & \ms{113.46}{0.19} & \ms{13.24}{0.02} \\
vMF      & 5  & \ms{104.55}{0.44} & \ms{116.84}{0.41} & \ms{12.29}{0.04} \\
Power    & 5  & \ms{106.09}{0.65} & \ms{118.20}{0.59} & \ms{12.12}{0.07} \\
\midrule
Gaussian & 10 & \ms{81.07}{0.23} & \ms{100.47}{0.14} & \ms{19.40}{0.17} \\
spCauchy & 10 & \bmsd{80.30}{0.38} & \ms{100.73}{0.34} & \ms{20.42}{0.06} \\
vMF      & 10 & \ms{83.38}{0.36} & \ms{102.97}{0.30} & \ms{19.59}{0.07} \\
Power    & 10 & \ms{84.12}{0.54} & \ms{103.70}{0.44} & \ms{19.58}{0.11} \\
\midrule
Gaussian & 20 & \ms{75.08}{0.27} & \ms{98.28}{0.18} & \ms{23.20}{0.12} \\
spCauchy & 20 & \bmsd{74.43}{0.30} & \ms{101.80}{0.19} & \ms{27.37}{0.15} \\
vMF      & 20 & \ms{97.25}{45.06} & \ms{118.82}{34.04} & \ms{21.57}{11.02} \\
Power    & 20 & \ms{82.29}{6.02} & \ms{108.07}{5.05} & \ms{25.78}{0.97} \\
\bottomrule
\end{tabular}
\end{table}

The paired improvements over the next-lowest method are $1.75$, $1.81$, $2.18$, $0.76$, and $0.65$ for increasing $p$. Their $95\%$ intervals are $[0.39,3.11]$, $[1.14,2.49]$, $[0.58,3.78]$, $[0.21,1.32]$, and $[0.36,0.94]$. The corresponding adjusted values are $0.0233$, $0.0086$, $0.0233$, $0.0233$, and $0.0086$.

All $100$ runs remain finite. At $p=20$, spherical Cauchy reconstruction ranges from $74.01$ to $74.77$, while Power Spherical ranges from $77.77$ to $92.63$ and vMF contains one collapsed value.

\begin{figure}[htbp]
    \centering
    \includegraphics[width=0.96\textwidth]{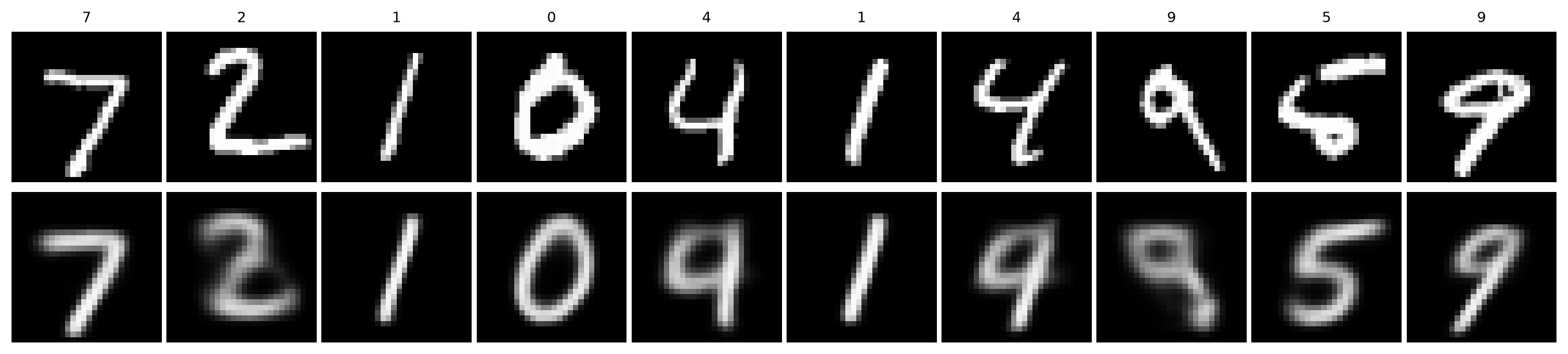}\\[3mm]
    \includegraphics[width=0.96\textwidth]{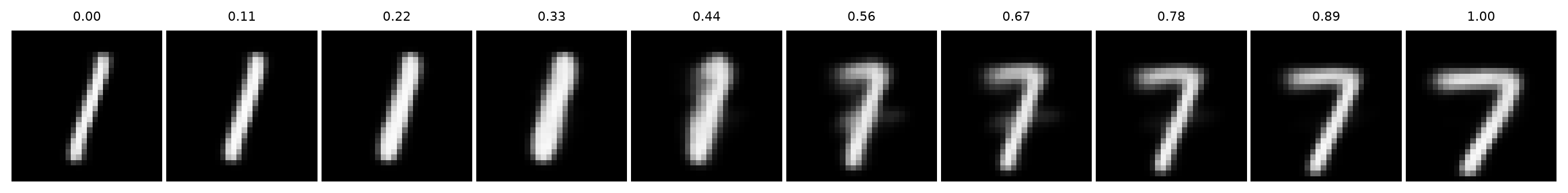}
    \caption{Representative MNIST reconstructions and a great-circle interpolation between encoded digits ``1'' and ``7'' for the selected spherical Cauchy run on $\mathbb S^2$.}
    \label{fig:supp-mnist-qualitative}
\end{figure}

\begin{figure}[htbp]
    \centering
    \includegraphics[width=\textwidth]{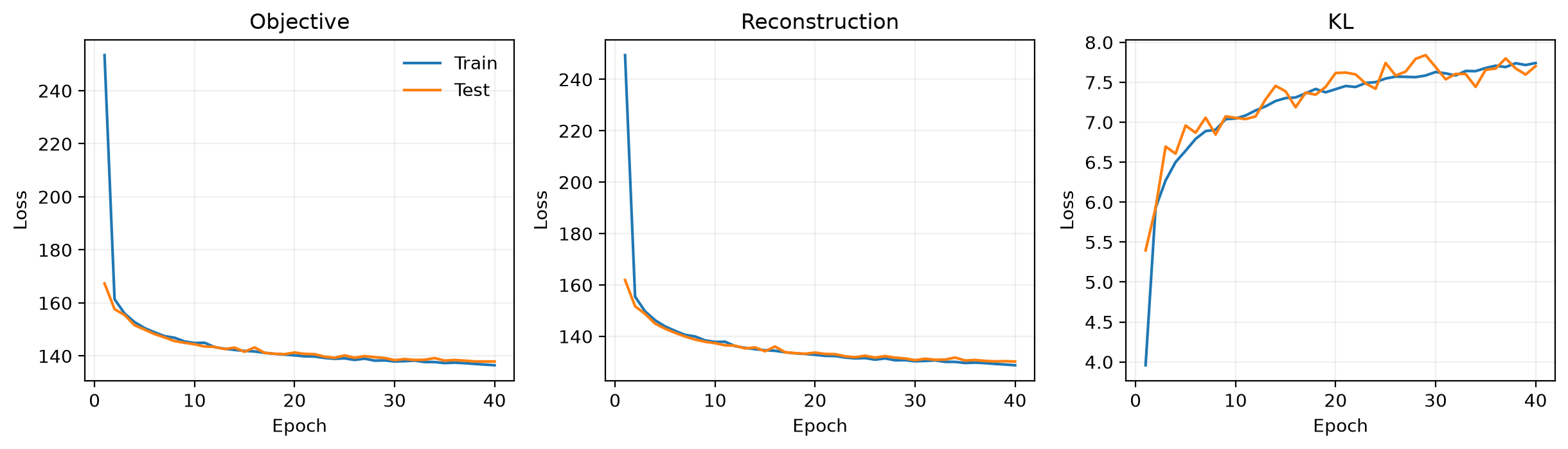}
    \caption{Training and evaluation objectives for the selected spherical Cauchy run on $\mathbb S^2$.}
    \label{fig:supp-mnist-convergence}
\end{figure}

\section{Controlled directional contamination}
\label{app:contamination}

The target in Equation~\eqref{eq:contaminated-target} uses $D=33$, $\kappa\in\{20,100,500\}$, and $\epsilon\in\{0,.01,.05,.10,.20\}$. Each family learns location and concentration from the same target samples. Fits use Adam for $600$ steps, batches of $8192$, and one million common held-out draws for evaluation. Three paired seeds change only the tangent direction of an initial $10^\circ$ location offset. The initialization and optimizer settings were fixed from diagnostic runs before the final grid.

\begin{table}[t]
\centering
\scriptsize
\setlength{\tabcolsep}{2.0pt}
\caption{Forward-KL results for the controlled contaminated target. Values are means and standard deviations over three seeds. Lower is better.}
\label{tab:supp-contaminated-forward}
\resizebox{\textwidth}{!}{%
\begin{tabular}{rlrrrrr}
\toprule
$\kappa$ & Model & $\epsilon=0$ & $0.01$ & $0.05$ & $0.10$ & $0.20$ \\
\midrule
20 & Spherical Cauchy & 0.0099 $\pm$ 0.0001 & \textbf{0.0125 $\pm$ 0.0000} & \textbf{0.0603 $\pm$ 0.0002} & \textbf{0.1396 $\pm$ 0.0004} & \textbf{0.2974 $\pm$ 0.0012} \\
 & vMF & \textbf{0.0000 $\pm$ 0.0000} & 0.0138 $\pm$ 0.0002 & 0.0993 $\pm$ 0.0002 & 0.2122 $\pm$ 0.0005 & 0.4014 $\pm$ 0.0013 \\
 & Power Spherical & 0.0105 $\pm$ 0.0001 & 0.0439 $\pm$ 0.0006 & 0.1975 $\pm$ 0.0004 & 0.3744 $\pm$ 0.0008 & 0.6340 $\pm$ 0.0018 \\
\addlinespace
100 & Spherical Cauchy & 0.0652 $\pm$ 0.0004 & \textbf{0.2137 $\pm$ 0.0007} & \textbf{0.8674 $\pm$ 0.0032} & \textbf{1.6811 $\pm$ 0.0007} & \textbf{3.1507 $\pm$ 0.0071} \\
 & vMF & \textbf{0.0001 $\pm$ 0.0000} & 0.5563 $\pm$ 0.0037 & 2.4400 $\pm$ 0.0069 & 4.1368 $\pm$ 0.0010 & 6.0941 $\pm$ 0.0065 \\
 & Power Spherical & 0.0014 $\pm$ 0.0001 & 0.8424 $\pm$ 0.0062 & 3.4991 $\pm$ 0.0074 & 5.6777 $\pm$ 0.0015 & 7.8836 $\pm$ 0.0063 \\
\addlinespace
500 & Spherical Cauchy & 0.0921 $\pm$ 0.0005 & \textbf{0.4759 $\pm$ 0.0016} & \textbf{2.0652 $\pm$ 0.0092} & \textbf{4.0321 $\pm$ 0.0030} & \textbf{7.6982 $\pm$ 0.0200} \\
 & vMF & \textbf{0.0001 $\pm$ 0.0000} & 3.6864 $\pm$ 0.0193 & 12.0475 $\pm$ 0.0278 & 16.8929 $\pm$ 0.0045 & 20.4315 $\pm$ 0.0082 \\
 & Power Spherical & 0.0002 $\pm$ 0.0000 & 5.0930 $\pm$ 0.0285 & 15.2234 $\pm$ 0.0253 & 20.4241 $\pm$ 0.0014 & 23.6803 $\pm$ 0.0047 \\
\bottomrule
\end{tabular}
}
\end{table}

\begin{table}[t]
\centering
\scriptsize
\setlength{\tabcolsep}{2.0pt}
\caption{Reverse-KL results for the controlled contaminated target. Values are means and standard deviations over three seeds. Lower is better.}
\label{tab:supp-contaminated-reverse}
\resizebox{\textwidth}{!}{%
\begin{tabular}{rlrrrrr}
\toprule
$\kappa$ & Model & $\epsilon=0$ & $0.01$ & $0.05$ & $0.10$ & $0.20$ \\
\midrule
20 & Spherical Cauchy & 0.0115 $\pm$ 0.0002 & 0.0101 $\pm$ 0.0001 & \textbf{0.0258 $\pm$ 0.0002} & \textbf{0.0549 $\pm$ 0.0003} & \textbf{0.1260 $\pm$ 0.0005} \\
 & vMF & \textbf{0.0000 $\pm$ 0.0000} & \textbf{0.0047 $\pm$ 0.0000} & 0.0306 $\pm$ 0.0000 & 0.0679 $\pm$ 0.0000 & 0.1519 $\pm$ 0.0001 \\
 & Power Spherical & 0.0090 $\pm$ 0.0001 & 0.0165 $\pm$ 0.0001 & 0.0489 $\pm$ 0.0001 & 0.0923 $\pm$ 0.0002 & 0.1872 $\pm$ 0.0003 \\
\addlinespace
100 & Spherical Cauchy & 0.1005 $\pm$ 0.0008 & 0.1084 $\pm$ 0.0007 & 0.1484 $\pm$ 0.0007 & 0.2017 $\pm$ 0.0007 & 0.3184 $\pm$ 0.0007 \\
 & vMF & \textbf{0.0001 $\pm$ 0.0000} & \textbf{0.0101 $\pm$ 0.0000} & \textbf{0.0514 $\pm$ 0.0000} & \textbf{0.1054 $\pm$ 0.0000} & \textbf{0.2232 $\pm$ 0.0000} \\
 & Power Spherical & 0.0014 $\pm$ 0.0000 & 0.0115 $\pm$ 0.0000 & 0.0527 $\pm$ 0.0000 & 0.1068 $\pm$ 0.0000 & 0.2246 $\pm$ 0.0000 \\
\addlinespace
500 & Spherical Cauchy & 0.1577 $\pm$ 0.0011 & 0.1677 $\pm$ 0.0011 & 0.2089 $\pm$ 0.0011 & 0.2629 $\pm$ 0.0011 & 0.3807 $\pm$ 0.0011 \\
 & vMF & \textbf{0.0001 $\pm$ 0.0000} & \textbf{0.0101 $\pm$ 0.0000} & \textbf{0.0514 $\pm$ 0.0000} & \textbf{0.1055 $\pm$ 0.0000} & \textbf{0.2232 $\pm$ 0.0000} \\
 & Power Spherical & 0.0002 $\pm$ 0.0000 & 0.0102 $\pm$ 0.0000 & 0.0515 $\pm$ 0.0000 & 0.1056 $\pm$ 0.0000 & 0.2233 $\pm$ 0.0000 \\
\bottomrule
\end{tabular}
}
\end{table}

At $\epsilon=0$, vMF wins every concentration under both objectives. Spherical Cauchy wins every positive-contamination forward-KL cell. Reverse KL favors the dominant vMF mode at $\kappa\in\{100,500\}$ and selects spherical Cauchy from $\epsilon=.05$ at $\kappa=20$.

\section{smallNORB experiment details}
\label{app:smallnorb}

We use the official instance partition and the left camera. Images are center-cropped from $96\times96$ to $80\times80$ and resized to $64\times64$. Instance $9$ in each category is reserved for validation. Removing azimuths $320^\circ$, $340^\circ$, $0^\circ$, and $20^\circ$ leaves $15{,}120$ training images and $4{,}860$ validation images. The official test set contains $5{,}400$ gap images and $18{,}900$ observed-viewpoint images.

All families use the same $6.35$ million-parameter CNN. The intrinsic latent dimension is $32$. Spherical models use $D=33$, while the parameter-matched Gaussian uses a $32$-dimensional mean and one learned isotropic scale. Training uses AdamW, batch size $128$, learning rate $10^{-4}$ with warm-up and cosine decay, weight decay $10^{-4}$, gradient clipping at $5$, and at most $100$ epochs. The KL weight increases from zero to $0.25$ over $30$ epochs. Checkpoints are selected by validation reconstruction NLL. A validation-only residual-CNN pilot performed worse, so the shared prescribed architecture was retained. The robust vMF path uses a half-integer Bessel recurrence specific to the tested $D=33$ configuration. The shared runtime and MNIST implementations are unchanged.

\begin{table}[t]
\centering
\scriptsize
\setlength{\tabcolsep}{2.0pt}
\caption{Complete smallNORB comparison over five paired seeds. Lower is better except for Spearman correlation.}
\label{tab:supp-smallnorb}
\begin{tabular}{lrrrrrr}
\toprule
Model & Gap NLL & Gap MSE & Observed NLL & Pose error & Spearman & Interp. MSE \\
\midrule
Gaussian & \ms{173.14}{2.77} & \ms{.00338}{.00005} & \ms{147.32}{2.29} & \ms{151.69}{.83} & \ms{-.054}{.009} & \ms{.00855}{.00006} \\
Power Spherical & \ms{130.11}{2.71} & \ms{.00254}{.00005} & \ms{112.44}{2.16} & \ms{148.11}{.95} & \bms{-.014}{.005} & \ms{.00839}{.00014} \\
Spherical Cauchy & \bmsd{122.90}{1.05} & \bmsd{.00240}{.00002} & \bmsd{106.18}{.88} & \bms{146.76}{.90} & \ms{-.018}{.005} & \ms{.00836}{.00012} \\
Robust vMF & \ms{127.55}{1.10} & \ms{.00249}{.00002} & \ms{110.34}{.48} & \ms{147.73}{1.00} & \ms{-.016}{.007} & \bms{.00825}{.00016} \\
\bottomrule
\end{tabular}
\end{table}

For gap NLL, spherical Cauchy beats robust vMF in all five paired seeds. The mean difference is $-4.6495$, its $95\%$ interval is $[-6.3682,-2.9307]$, and the paired two-sided $t$-test gives $p=.00168$. The paired comparison with Power Spherical is $-7.2067$, with interval $[-10.8186,-3.5949]$ and $p=.00519$. Pose, distance, and interpolation summaries are reported in
Table~\ref{tab:supp-smallnorb}, factor and training diagnostics are shown below.

\begin{figure}[htbp]
    \centering
    \includegraphics[width=\textwidth]{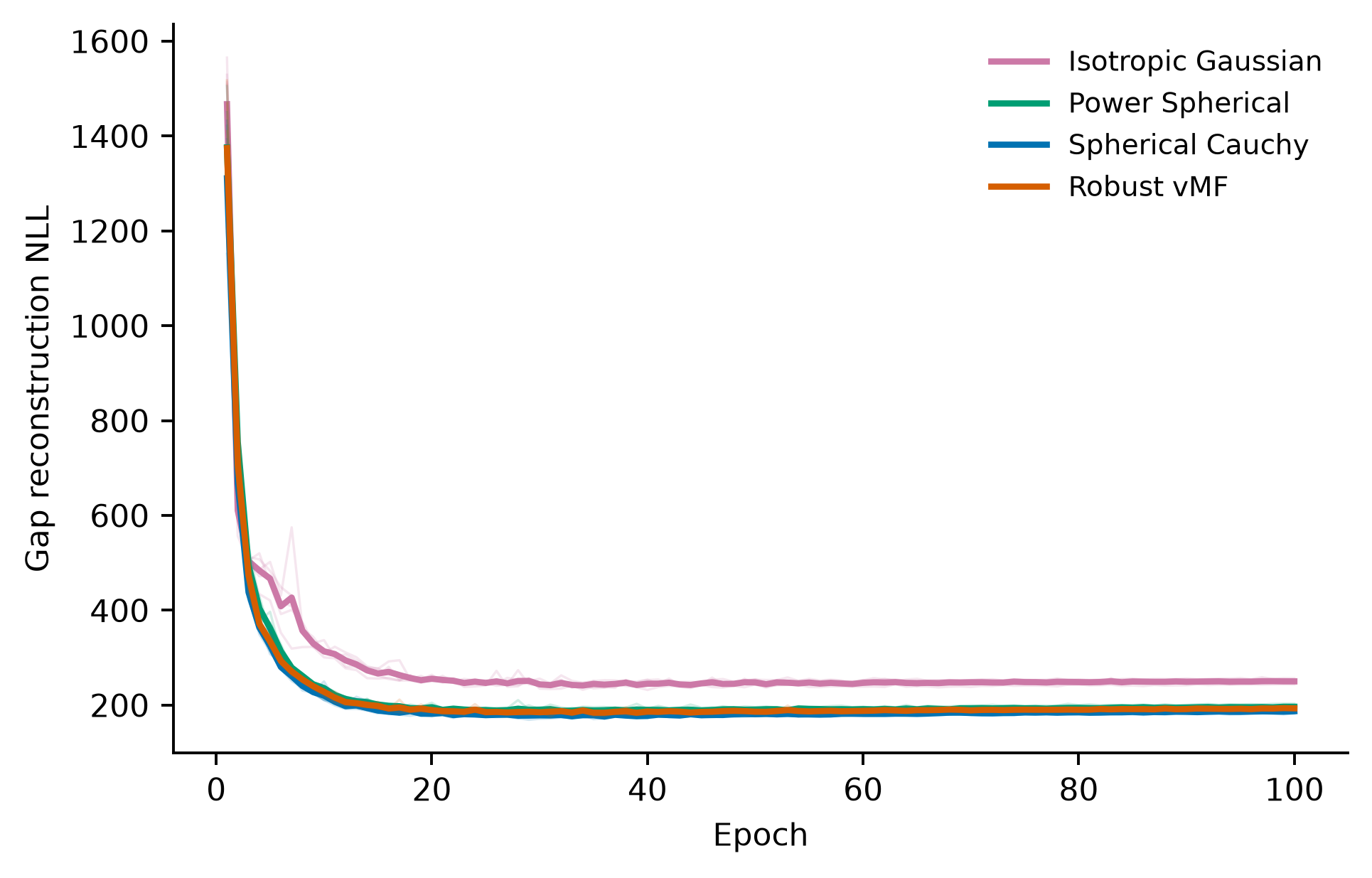}
    \caption{smallNORB validation-gap reconstruction trajectories over five seeds.}
    \label{fig:supp-smallnorb-training}
\end{figure}

\begin{figure}[htbp]
    \centering
    \includegraphics[width=.49\textwidth]{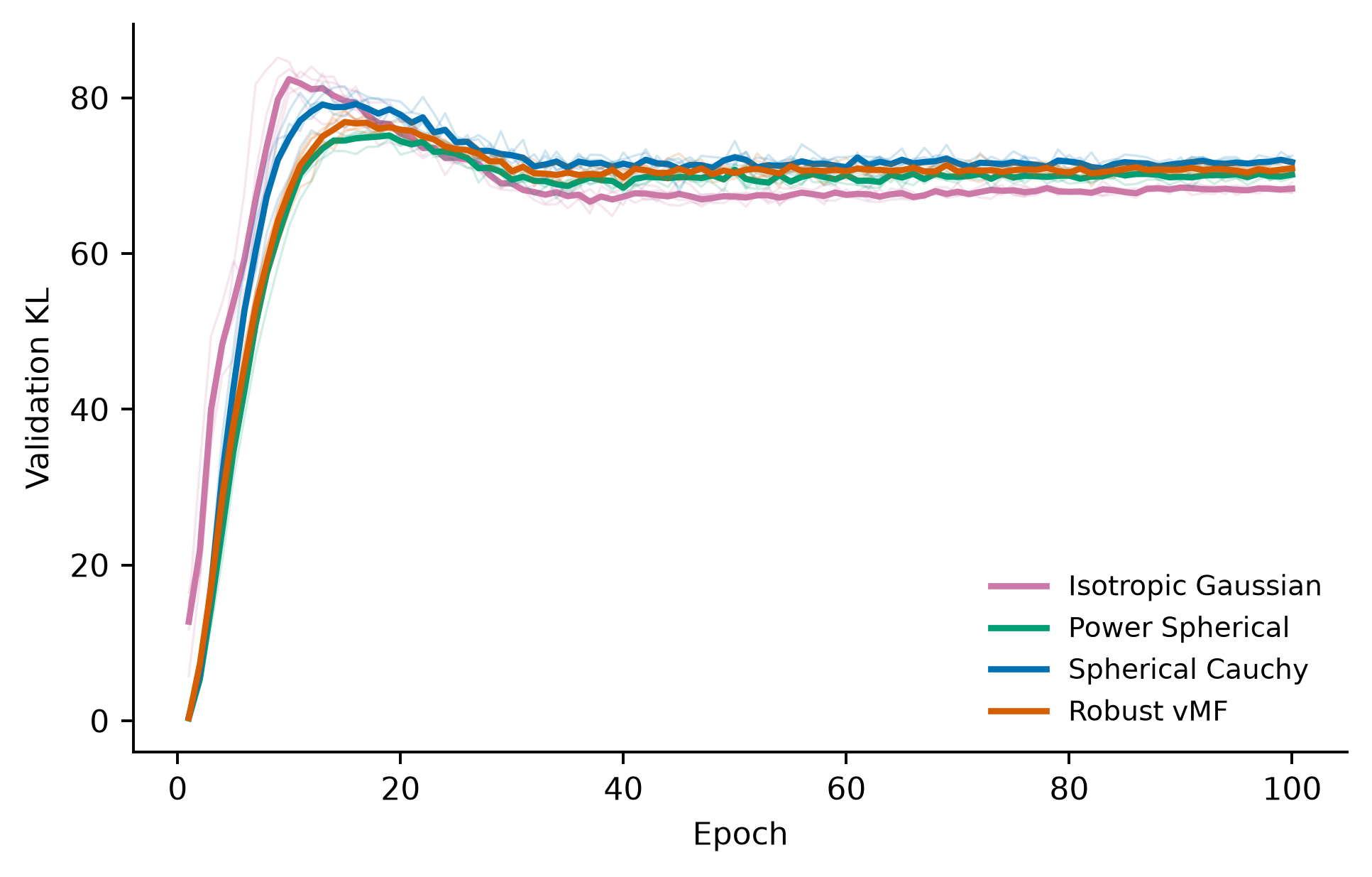}\hfill
    \includegraphics[width=.49\textwidth]{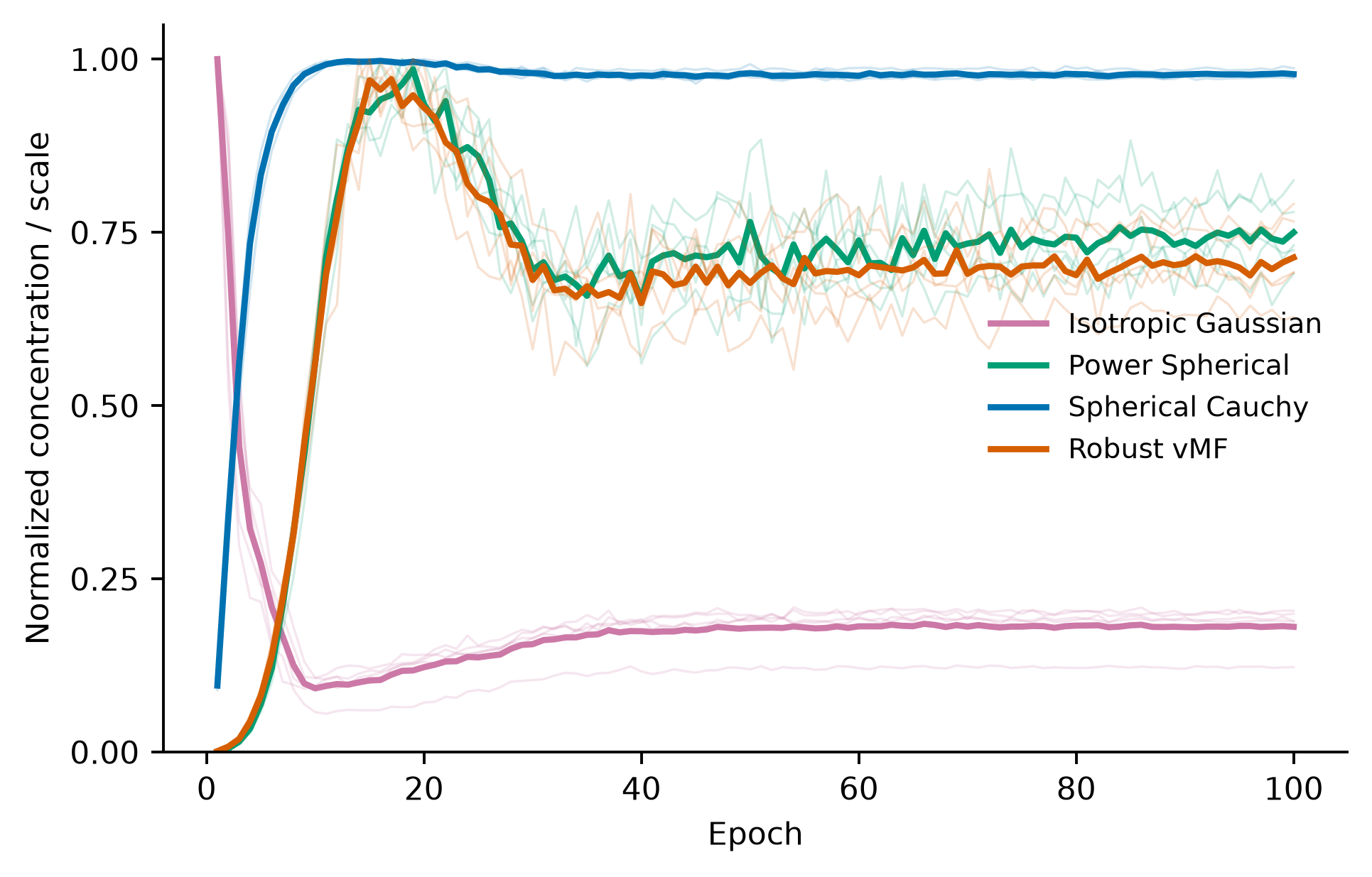}
    \caption{Posterior KL (left) and family-specific posterior concentration/scale trajectories (right) during smallNORB training. Each right-panel seed trajectory is divided by its own maximum over training before the five-seed average is computed.}
    \label{fig:supp-smallnorb-posterior}
\end{figure}

\begin{figure}[htbp]
    \centering
    \includegraphics[width=\textwidth]{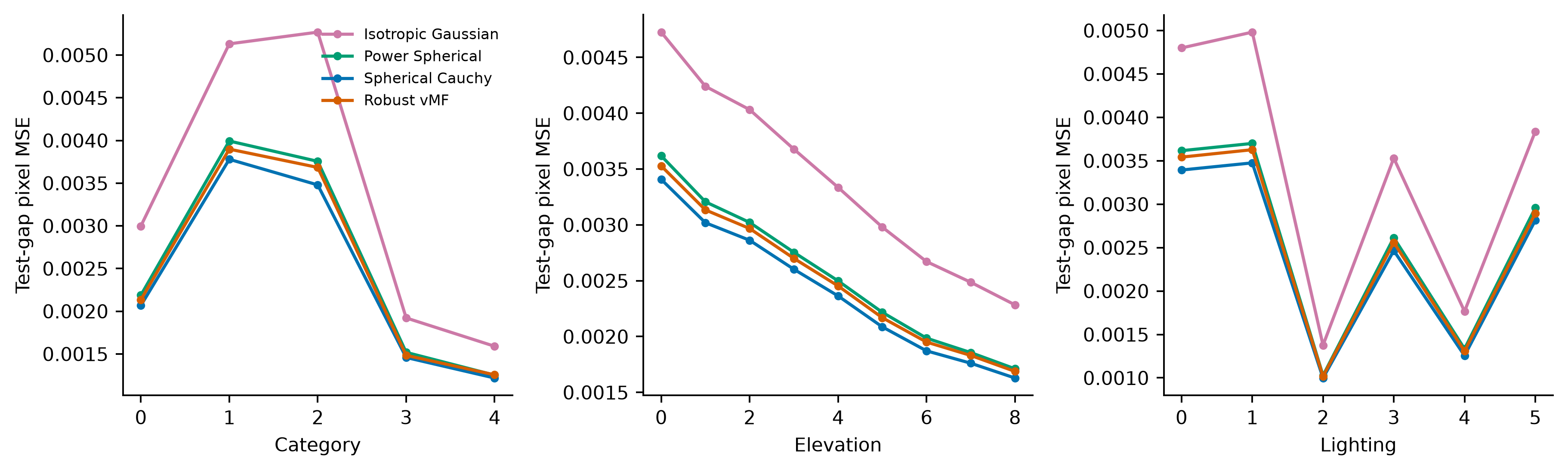}
    \caption{smallNORB reconstruction separated by category, elevation, and lighting.}
    \label{fig:supp-smallnorb-factors}
\end{figure}

\end{document}